\documentclass[lettersize,journal]{IEEEtran}
\usepackage{amsmath,amsfonts}
\usepackage{algorithmic}
\usepackage{algorithm}
\usepackage{amsmath}

\usepackage{array}
\usepackage[caption=false,font=normalsize,labelfont=sf,textfont=sf]{subfig}
\usepackage{textcomp}
\usepackage{stfloats}
\usepackage{url}
\usepackage{verbatim}
\usepackage{graphicx}
\usepackage{cite}
\usepackage{subcaption}
\usepackage{subfig}
\usepackage{xcolor}  % Required for colored text
\definecolor{darkgreen}{rgb}{0.0, 0.6, 0.0} % Deep green
\definecolor{darkred}{rgb}{0.8, 0.0, 0.0}   % Deep red
\definecolor{yellow}{rgb}{0.9, 0.8, 0.0}    % Yellow for medium
\begin{document}

\title{MeshONet: A Generalizable and Efficient Operator Learning Method for Structured Mesh Generation}

\author{
    Jing Xiao, % 第一作者
    Xinhai Chen\IEEEauthorrefmark{1}, % 第二作者，标记为通讯作者,
    Jiaming Peng,
    Qingling Wang,
    Jie Liu% 第三作者
    \IEEEauthorblockA{
        \\
        Laboratory of Digitizing Software for Frontier Equipment, \\
        Science and Technology on Parallel and Distributed Processing Laboratory,\\
        National University of Defense Technology, Changsha 410073.
    }\\
    \thanks{\IEEEauthorrefmark{1}Xinhai Chen, Corresponding Author, Email: chenxinhai16@nudt.edu.cn}
}

\maketitle

\begin{abstract}
Mesh generation plays a crucial role in scientific computing. Traditional mesh generation methods, such as TFI and PDE-based methods, often struggle to achieve a balance between efficiency and mesh quality. To address this challenge, physics-informed intelligent learning methods have recently emerged, significantly improving generation efficiency while maintaining high mesh quality. However, physics-informed methods fail to generalize when applied to previously unseen geometries, as even small changes in the boundary shape necessitate burdensome retraining to adapt to new geometric variations. In this paper, we introduce MeshONet, the first generalizable intelligent learning method for structured mesh generation. The method transforms the mesh generation task into an operator learning problem with multiple input and solution functions. To effectively overcome the multivariable mapping restriction of operator learning methods, we propose a dual-branch, shared-trunk architecture to approximate the mapping between function spaces based on input-output pairs. Experimental results show that MeshONet achieves a speedup of up to four orders of magnitude in generation efficiency over traditional methods. It also enables generalization to different geometries without retraining, greatly enhancing the practicality of intelligent methods.
\end{abstract}

\begin{IEEEkeywords}
Structured Mesh Generation, Neural network, Operator Learning, Generalization
\end{IEEEkeywords}

\begin{table*}[h!]
    \centering
    \caption{This table compares various mesh generation methods. A checkmark `\textcolor{darkgreen}{$\checkmark$}' indicates strong performance, and a `\textcolor{darkred}{$\times$}' denotes poor performance.}

    \renewcommand{\arraystretch}{1.5} % Adjust row height
    \begin{tabular}{@{\hskip 0pt}>{\centering\arraybackslash}p{3.5cm} @{\hskip 0pt}>{\centering\arraybackslash}p{2.5cm} @{\hskip 0pt}>{\centering\arraybackslash}p{2.5cm} @{\hskip 0pt}>{\centering\arraybackslash}p{2.5cm} @{\hskip 0pt}>{\centering\arraybackslash}p{3cm}@{\hskip 0pt}}
        \hline
        \multicolumn{1}{c}{Method} & \multicolumn{1}{c}{TFI} & \multicolumn{1}{c}{PDE-based} & \multicolumn{1}{c}{Physics-Informed} & \multicolumn{1}{c}{\textbf{MeshONet (Ours)}} \\
        \hline
        Generation Efficiency & \textcolor{darkgreen}{$\checkmark$} & \textcolor{darkred}{$\times$} & \textcolor{darkgreen}{$\checkmark$} & \textcolor{darkgreen}{$\checkmark$} \\
        Mesh Quality &  \textcolor{darkred}{$\times$} &  \textcolor{darkgreen}{$\checkmark$} & \textcolor{darkgreen}{$\checkmark$} & \textcolor{darkgreen}{$\checkmark$}   \\
        \textbf{Generalization} & \textcolor{darkgreen}{$\checkmark$} & \textcolor{darkgreen}{$\checkmark$} & \textcolor{darkred}{$\times$} & \textcolor{darkgreen}{$\checkmark$} \\
        \hline
    \end{tabular}
   
    \label{tab:mesh_methods}
\end{table*}

\section{Introduction}
\begin{figure}[t]
	\centering
		\includegraphics[width = 0.5\textwidth]{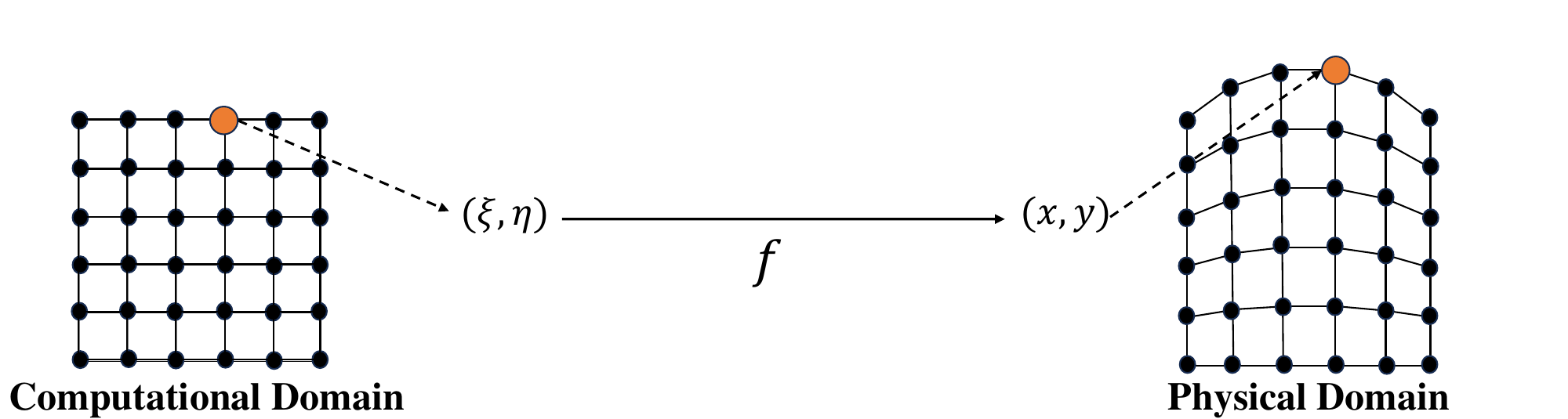}
	\caption{Schematic of the structured mesh generation process, showing the mapping of the computational mesh \((\xi, \eta)\) to the physical mesh \((x, y)\) through the function \( f \).}
	\label{FIG:网格生成示意图}
\end{figure}

\IEEEPARstart{M}{esh} generation is extensively utilized across diverse scientific and engineering domains, such as aerospace engineering \cite{blockley2010encyclopedia}, automotive engineering \cite{crolla2015encyclopedia}, meteorology \cite{petterssen2011introduction}, ocean engineering \cite{mccormick2009ocean}, and materials science \cite{callister2020materials}. As a crucial preprocessing step, mesh generation involves discretizing the continuous physical domain into a finite set of cells, facilitating the application of numerical methods for analysis. Among various types of meshes, structured meshes are widely employed in practical applications due to their regular indexing, topological connectivity, and superior computational efficiency. The core of the structured mesh generation process lies in solving the mapping between the computational domain and the physical domain. This process can be expressed as a one-to-one mapping from computational coordinates \( (\xi, \eta) \) to physical coordinates \( (x, y) \), as illustrated in Figure \ref{FIG:网格生成示意图}.

Traditional methods for solving the mapping from the computational domain to the physical domain are primarily represented by transfinite interpolation (TFI) and partial differential equation-based (PDE-based) approaches. These methods have been widely utilized in structured mesh generation. While these  methods have proven effective in many scenarios, they often face challenges in balancing mesh quality and generation efficiency. TFI can quickly generate structured meshes and is computationally simple and efficient \cite{rvachev2001transfinite}; however, they tend to produce degenerate, overlapping, or boundary-crossing elements in complex geometries, leading to a decline in mesh quality.
 PDE-based methods consider the mapping solving problem as a boundary value problem of solving PDEs that govern the distribution of mesh points and their boundary conditions. By iteratively solving the corresponding equations, they generate high-quality structured meshes, making them particularly suitable for complex geometric structures \cite{babuska2012modeling}. However, PDE-based methods incur substantial computational costs, especially in large-scale mesh generation tasks, where the process can be exceedingly slow and may not meet the demands of real-time or large-scale applications.
To address the limitations of traditional methods, physics-informed methods have emerged as a promising solution \cite{chen2022mgnet,peng20243dmeshnet}. These methods leverage the universal approximation capability of deep neural networks to capture the input-output relationship of the domain in high-dimensional nonlinear space. Consequently, these methods eliminate the need for explicit PDE iterative solving, enabling fast feedforward prediction and thus significantly improving computational efficiency. Furthermore, by incorporating the control equations and boundary conditions related to mesh generation into the neural network loss function, physics-informed methods effectively guide the mesh generation process to adhere to physical geometric constraints, thereby ensuring the generation of high-quality meshes. One significant drawback of physics-informed methods is their limited generalization to previously unseen geometries. When the geometries change, the corresponding boundary conditions will be modified, requiring the loss function to be readjusted. As a result, even small changes in boundary shapes require burdensome retraining to adapt to new geometric variations.

\begin{figure}[t]
	\centering
\includegraphics[width=0.5\textwidth,height=8cm]{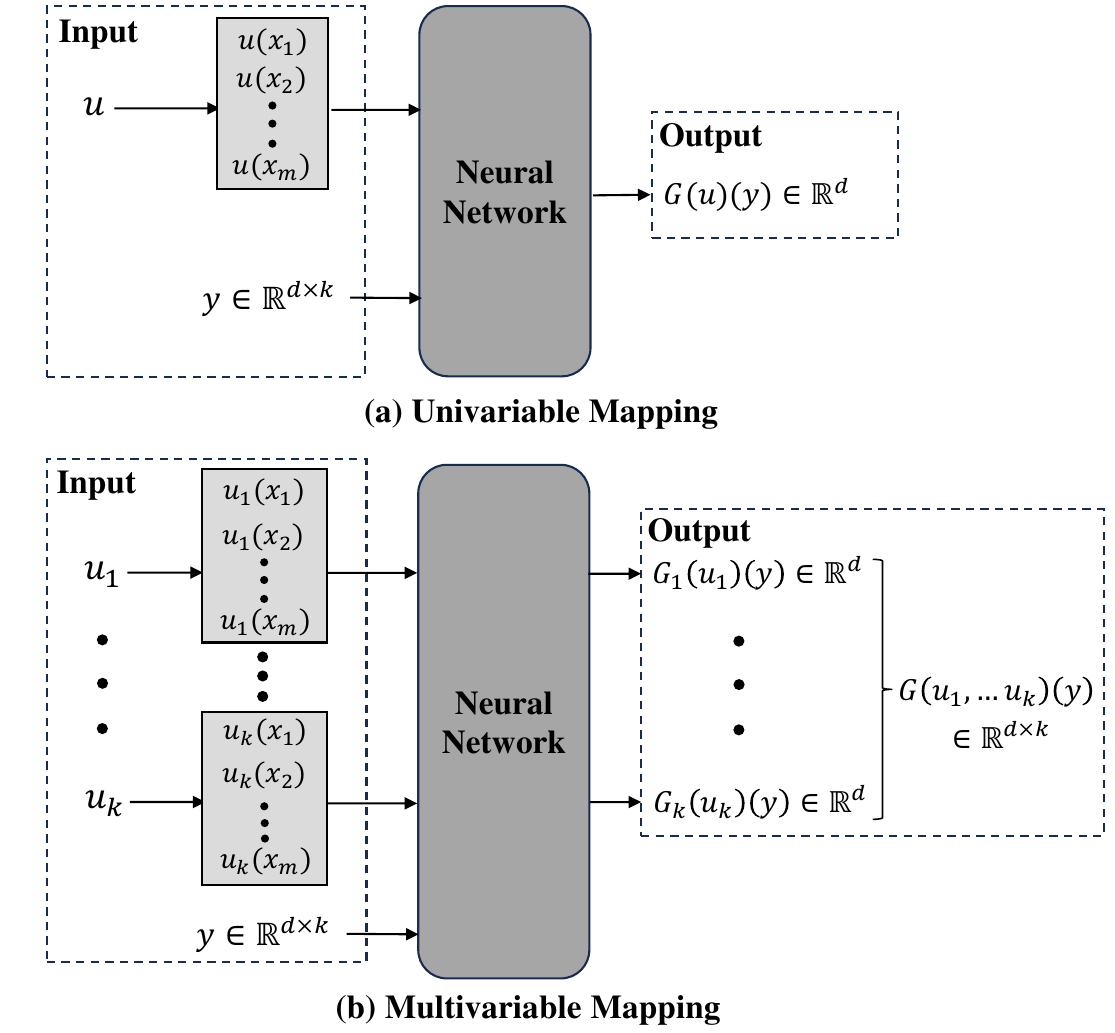}
	\caption{Illustration of univariable and multivariable mapping. The input set $\{x_1, x_2, \dots, x_m\}$ represents the sensor locations, where each $x_i \in \mathbb{R}^k$.  The variable \( y \) represents the evaluation points, with \( d \) points, each having \( k \) dimensions. \( G \) is the operator that maps the input functions to their corresponding solution values.
 \textbf{(a) Univariable Mapping:} The input is a single function \( u \), evaluated at \( m \) sensors, and the output is a \( d \)-dimensional vector \( G(u)(y) \), representing the solution values at \( d \) evaluation points.  \textbf{(b) Multivariable Mapping:} 
  The input consists of \( k \) functions \( u_1, u_2, \ldots, u_k \), evaluated at \( m \) sensors. The output is a \( d \times k \) matrix \( G(u_1, \ldots, u_k)(y) \), where each row represents the solution values at \( d \) evaluation points \( y \), and the \( j \)-th column, generated by the sub-operator \( G_j \), represents the \( j \)-th component of the solution, where \( j = 1, \dots, k \).
}
	\label{FIG:多输入多输出示意图}
\end{figure}

As illustrated in Table \ref{tab:mesh_methods}, both traditional and physics-informed methods exhibit notable limitations. In response, operator learning-based methods offer a fresh perspective to address the limitations of established approaches. The goal of operator learning is to learn the generalizable mapping between function spaces based on input-output pairs \cite{effros2022theory}. By learning this mapping, it enables direct prediction of the solution from different input conditions, without the need for explicit equation solving and retraining, thereby significantly improving computational efficiency and enhancing generalization capability. However existing operator learning-based methods are limited to univariable mapping problems, which involve a single input function and a single output function. In 2D mesh generation, the task typically involves two boundary functions as inputs  and two corresponding solution functions, thereby forming a multivariable mapping problem. This distinction between univariable and multivariable mapping problems, as illustrated in Figure \ref{FIG:多输入多输出示意图}, makes them unsuitable for structured mesh generation.
 Therefore, effectively leveraging the generalization capabilities of operator learning while addressing the challenges of multivariable mapping remains an open problem.

The main contributions of this work are as follows:
\begin{itemize}
    \item We propose MeshONet, a generalizable and efficient structured mesh generation method, which transforms the mesh generation task into an operator learning problem to capture the underlying meshing rules.

    \item We propose an operator learning-based architecture specifically designed to address the multivariable mapping problem in mesh generation tasks. The network employs a dual-branch, shared-trunk structure and incorporates sub-sampling and random sampling strategies to enhance its performance.

    \item We conduct a comprehensive evaluation of MeshONet across a range of test cases, showcasing its ability to generalize across different geometries without retraining. It achieves up to a four-order-of-magnitude improvement in mesh generation efficiency, while maintaining high mesh quality.
\end{itemize}

To the best of our knowledge, we are the first to apply operator learning to structured mesh generation. This approach provides a reference framework for future applications involving multivariable mapping problems. The paper is organized as follows: Section II reviews related work, Section III details our proposed method, Section IV presents experimental results, and Section V concludes with a summary and future directions.

\section{Related Work}
Mesh generation is a critical component in scientific computing. Traditional methods, such as algebraic and PDE-based approaches, have been widely used. Algebraic mesh generation relies on geometric interpolation to generate meshes. A representative method in this category is TFI \cite{nguyen1989applications}, which rapidly generates structured meshes by interpolating boundary conditions. The TFI method constructs internal meshes by interpolating known boundaries, making it suitable for regular geometric structures and simple boundary conditions. Due to the relatively simple computation steps, algebraic methods have a fast generation speed. However, algebraic methods face considerable limitations when applied to intricate geometries, such as airfoils or shapes with high curvature. Their interpolation processes are inadequate for handling high-curvature and irregular geometries, often resulting in low-quality mesh elements such as degenerate triangles, overlapping elements, or meshes that cross boundaries. 

To address the boundary issues of algebraic mesh generation, PDE-based methods are often used for structured mesh generation tasks with complex boundary conditions. This methods generate the corresponding mesh by solving PDEs in the given transformation domain, thereby linking computational domain coordinates to physical domain coordinates and yielding the steady-state solution as the final structured mesh.
Based on the mathematical properties of PDEs, PDE-based methods can ensure orthogonality in near-wall regions, prevent the propagation of boundary singularities, and avoid overlapping mesh cells. These capabilities enable the generation of high-quality meshes, making them widely used in high-precision numerical simulations, such as turbulence simulation in fluid mechanics \cite{moin1998direct} or atmospheric boundary layer calculations in aerospace applications \cite{kaimal1994atmospheric}. However, the computational complexity of PDE-based methods is high; the process requires iterative solutions to control equations to generate meshes, resulting in substantial computational costs and slower generation speeds, especially in large-scale mesh generation tasks. Therefore, achieving a balance between mesh generation efficiency and quality is a significant challenge for traditional methods.

To overcome the limitations of traditional methods, researchers have explored the use of neural networks for mesh generation. Pioneering works in this area include those of Alfonzetti et al. \cite{2002Automatic}, who were among the first to apply neural networks to unstructured mesh generation. They began with coarse triangular meshes containing a limited number of nodes and progressively refined the mesh by adding points until the desired resolution was reached. Similarly, Ahmet et al. \cite{ahmet2002neural} used boundary points as inputs to a feedforward, single-layer neural network, which predicted the spatial coordinates of interior points, enabling 2D mesh generation. But these methods were highly sensitive to weight selection, with improper choices leading to significant inaccuracies in the meshes.

In order to improve the robustness of the model, several researchers have started to investigate data-driven approaches for mesh generation.
 Lowther et al. \cite{lowther1993density} use density information to optimize element size and placement, offering more efficient mesh generation compared to traditional adaptive methods. Besides, Zhang et al. \cite{zhang2020meshingnet} introduced a novel deep learning-based method to automatic unstructured mesh generation, where they used a finite element solver to perform simulations, thereby creating a training dataset. Papagiannopoulos et al. \cite{papagiannopoulos2021teach} use data extracted from meshed contours to train neural networks, which are then employed to approximate the number of vertices to be inserted inside the contour cavity, along with their location and connectivity. Nevertheless, these methods typically exhibit weak extrapolation capabilities and require large amounts of data, which limits their applicability in real-world scenarios.

In response to the challenge of expensive data acquisition in data-driven methods, physics-informed methods have emerged, which incorporate governing equations and boundary conditions as physical constraints in the loss function, thereby effectively constraining the solution space and eliminating the need for prior data. In this regard, MGNet \cite{chen2022mgnet} is a pioneering work, being the first to use a physics-informed neural network \cite{raissi2019physics,kapoor2023physics,hua2023physics} for structured mesh generation.  By minimizing the weighted residuals of the control equations and boundary constraints, MGNet effectively learns an approximate mapping for mesh generation, significantly improving both efficiency and accuracy. Building on this, Chen et al. further advanced the field with MeshNet \cite{chen2023developing}, a method that leverages  data from coarse mesh generation to optimize the loss function, thereby improving both the quality and efficiency of mesh generation. Additionally, they introduced another approach that incorporates data from auxiliary line sampling to construct a more refined loss function, further enhancing the performance of mesh generation \cite{chen2022improved}. Besides, research in physics-informed methods has expanded to three-dimensional scenarios. 3DMeshNet \cite{peng20243dmeshnet} is a representative work in this area. This method improves adaptability to complex geometric structures by adjusting weights and projecting gradients within the loss function. It also simplifies derivative calculations using finite difference methods, thereby enabling the efficient generation of three-dimensional structured meshes. However, these physics-informed methods are limited in their generalization, especially with unseen geometries. 
When geometries change, the boundary conditions also change, which in turn necessitates readjusting the loss function. Consequently, even minor changes in boundary shapes require burdensome retraining to accommodate new variations.

With the aim of addressing the challenge posed by the limited generalization capabilities of physics-informed methods, operator learning-based methods provide a promising solution.  By learning the nonlinear mapping between inputs and outputs, operator learning effectively infers the behavior of complex methods from limited training data, showcasing exceptional generalization capabilities \cite{lu2021learning,kovachki2023neural,li2020fourier,tripura2023wavelet,xiong2024koopman,cao2023lno}. The generalization capabilities of operator learning suggest its potential for mesh generation. By learning from given input-output pairs, operator learning-based methods can learn the operator that captures the underlying relationship between the input and output spaces, thereby acquiring the ability to generalize. Existing operator learning-based methods can only handle univariable mapping problems. In contrast, mesh generation tasks require addressing multivariable mapping problems. This difference limits the application of operator learning to mesh generation.

In this paper, we propose MeshONet to address the limitations of the aforementioned methods. We demonstrate that MeshONet improves generation efficiency while maintaining mesh quality. It also generalizes across different geometries and effectively addresses multivariable mapping problems.

\vspace{2cm}
\section{MeshONet:A Generalizable and Efficient Structured Mesh Generation
method}

\subsection{Problem Setting}
The structured mesh generation process can be described as solving a boundary value problem. An elliptic partial differential equation is taken as an example:
\begin{equation} 
\begin{aligned} 
&\alpha x_{\xi \xi} - 2\beta x_{\xi,\eta} + \gamma x_{\eta \eta} = 0, \quad (\xi,\eta) \in \Omega, \\
&\alpha y_{\xi \xi} - 2\beta y_{\xi,\eta} + \gamma y_{\eta \eta} = 0, \quad (\xi,\eta) \in \Omega, \\
&x = u_1(\xi,\eta), \quad (\xi,\eta) \in \partial\Omega, \\
&y = u_2(\xi,\eta), \quad (\xi,\eta) \in \partial\Omega. 
\end{aligned}
\end{equation} 
The coefficients $\alpha$, $\beta$, and $\gamma$ are defined as follows:
\begin{equation}
\begin{aligned}
&\alpha = x_{\eta}^{2} + y_{\eta}^{2}, \\
&\beta = x_{\xi} x_{\eta} + y_{\xi} y_{\eta}, \\
&\gamma = x_{\xi}^{2} + y_{\xi}^{2}.
\end{aligned}
\end{equation}
In this set of equations, \( \xi \) and \( \eta \) are the coordinates defined in the computational domain, while \( x \) and \( y \) are the coordinates defined in the physical domain. By solving the above elliptic differential equations, we can obtain the corresponding values of \( x = x(\xi, \eta) \) and \( y = y(\xi, \eta) \), which represent the mesh coordinates in the physical domain.

 Our objective is to learn the operator \( G \) that maps a Banach space \( \mathcal{U} \times \mathcal{U} \) to another Banach space \( \mathcal{P} \times \mathcal{P}\), where both \( u_1 \in \mathcal{U} \) and \( u_2 \in \mathcal{U} \) represent boundary conditions. This can be formally expressed as:  
\begin{equation}  
(u_1, u_2)  \mapsto G(u_1, u_2)    
\end{equation}

The operator \( G \) takes three inputs: the computational domain coordinates \( (\xi, \eta) \), and two boundary functions \( u_1 \) and \( u_2 \), which describe the \( x \)- and \( y \)-coordinates of the boundary, respectively.  For any given boundary conditions \( u_1 \) and \( u_2 \), and a set of computational domain coordinates \( (\xi, \eta) \), the corresponding physical mesh coordinates \( (x, y) \) are generated by the operator \( G \), as expressed by the equation:
\begin{equation}
    (x, y) = G(u_1, u_2)(\xi, \eta)
\end{equation}
The operator \( G \) can be decomposed into two distinct sub-operators: \( G_x \), responsible for generating the \( x \)-coordinates, and \( G_y \), responsible for generating the \( y \)-coordinates. Each of these sub-operators operates from the Banach space \( \mathcal{U} \) to \( \mathcal{P} \), where \( u_1 \mapsto G_x(u_1) \) and \( u_2 \mapsto G_y(u_2) \). These sub-operators are not independent, as they both rely on the same input coordinates \( (\xi, \eta) \). Consequently, the operator \( G(u_1, u_2) \) can be formally expressed as:
\begin{equation}
G(u_1, u_2)(\xi, \eta) = \left( G_x(u_1)(\xi, \eta), G_y(u_2)(\xi, \eta) \right)
\end{equation}

Inspired by the universal approximation theorem for continuous multi-input operators \cite{jin2022mionet}, which states that there exists a continuous multi-input operator that can be approximated by a neural network to arbitrary precision, we approximate the operator \( G(u_1, u_2) \) using a neural network-based approximation \( \hat{G}(u_1, u_2) \). The approximation is constructed to satisfy the following condition:
\begin{equation}
\| G(u_1, u_2)(\xi, \eta) - \hat{G}(u_1, u_2)(\xi, \eta) \| \to 0
\end{equation}
This provides a theoretical foundation for our method.

\begin{figure*}
	\centering
		\includegraphics[scale=0.5]{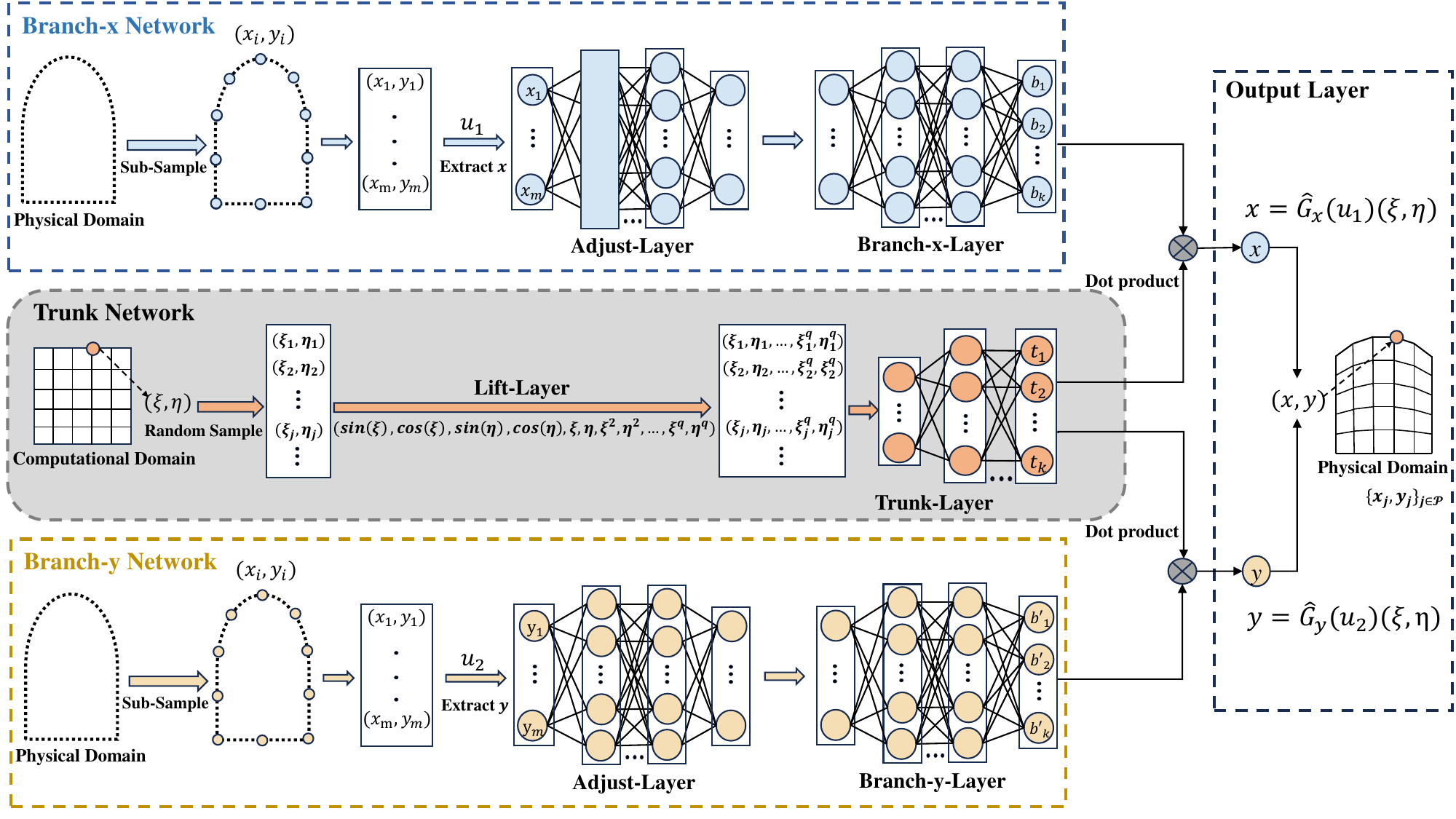}	\caption{Architecture of MeshONet for mesh generation. The model consists of three main components: Branch-x Network, Trunk Network, and Branch-y Network. The Branch-x and Branch-y networks process sub-sampled boundary data, extracting features specific to the \(x\) and \(y\) coordinates, respectively. The Trunk Network operates on the computational mesh, applying a Lift-Layer to transform input coordinates \((\xi, \eta)\) into a higher-dimensional space, followed by feature extraction in the Trunk-Layer. Outputs from the Branch and Trunk networks are combined using a dot product operation in the Output Layer, producing physical mesh coordinates \((x, y)\) as functions \(\hat{G}_x(u_1)(\xi, \eta)\) and \(\hat{G}_y(u_2)(\xi, \eta)\), effectively mapping the computational domain to the physical domain.}
	\label{FIG:MeshONet}
\end{figure*}

\subsection{The Operator Learning-based method}
As mentioned above, existing operator-based methods cannot be directly applied to mesh generation. Therefore, we propose an operator learning-based framework specifically designed for mesh generation, aimed at solving the multivariable mapping problem.

\subsubsection{The Architecture Of MeshONet}
The architecture of MeshONet consists of three components: two specialized branch networks, the Branch-x Network and the Branch-y Network, and a shared trunk network.

The 2D mesh generation task involves two boundary functions at each sensor, corresponding to the \(x\)-coordinates and \(y\)-coordinates of the boundary. Handling both functions simultaneously can be challenging. To address this, we employ a dual-branch network design, ensuring that each branch processes only one boundary function. Specifically, the Branch-x Network extracts the \(x\)-coordinates of the boundary points, while the Branch-y Network extracts the \(y\)-coordinates. Both networks take the boundary point coordinates as input. 

However, since there are infinitely many boundary points and the boundary conditions correspond to infinitely many values, this problem is inherently infinite-dimensional. To further address this, we employ a subsampling strategy in which fixed sensor locations in the computational domain are used to obtain the corresponding boundary points \((x_1, y_1), (x_2, y_2), \dots, (x_m, y_m)\) in the physical domain. These points are subsequently utilized to sample the boundary functions \(u_1\) and \(u_2\), where \(u_1\) and \(u_2\) represent the sampled values of the \(x\)-coordinates and \(y\)-coordinates, respectively.
 This results in two finite-dimensional representations, \([x_1, x_2, \dots, x_m]\) and \([y_1, y_2, \dots, y_m]\).

After obtaining the finite-dimensional representation of the boundary conditions, we use this feature representation as the input to the Adjust Layer.
The Adjust Layer consists of fully connected layers with Tanh activation functions. Specifically, the Adjust Layer is defined as:
\begin{equation}
\text{Adjust}(x) = f_{L}(f_{L-1}(...f_{1}((x)...)),
\end{equation}
where each function \( f_{i} \) represents the mapping of the \( i \)-th layer of the Adjust Layer, defined as:
\begin{equation}
f_{i}(x) = \tanh(W_{i}x + b_{i}),
\end{equation}
where \( W_{i} \) and \( b_{i} \) are the weight matrix and bias term of the \( i \)-th layer, respectively. Similarly, the Branch-y Network processes the \( y \)-coordinates in the same manner, forwarding them through the Adjust Layer before passing them to the corresponding Branch layer.

As the solutions for \( x \) and \( y \) in the structured mesh generation task are inherently coupled, we use a shared Trunk Network that takes the computational domain coordinates as input.
This network serves as a unified feature extractor for both the Branch-x and Branch-y Networks, ensuring the correlation between the \( x \)- and \( y \)-coordinates and promoting efficient information flow between the two branches, thus improving MeshONet's performance. In addition, we design a Lift-Layer that expands the computational domain coordinates \( (\xi, \eta) \) into a higher-dimensional space using a mixed dimensionality expansion approach. This layer combines polynomial and trigonometric transformations to capture linear, non-linear, and periodic features, thereby enhancing feature representation. The dimensionality expansion is given by:
\begin{equation}
(\xi, \eta) \mapsto [\sin(\xi), \cos(\xi), \sin(\eta), \cos(\eta), \xi, \eta, \xi^2, \eta^2, \dots, \xi^q, \eta^q].
\end{equation}

After obtaining the features extracted by the Branch and Trunk networks, they are fused using a dot product to generate the final coordinates. The generation process for the \( x \) and \( y \) coordinates is as follows:
\begin{equation}
x = \hat{G}_x(u_1)(\xi, \eta) = \text{Branch}_x(u_1) \cdot \text{Trunk}(\xi, \eta) = \sum_{i=1}^{k} b_{i} t_{i} + b_{0}.
\end{equation}
\begin{equation}
y = \hat{G}_y(u_2)(\xi, \eta) = \text{Branch}_y(u_2) \cdot \text{Trunk}(\xi, \eta) = \sum_{i=1}^{k} b'_{i} t_{i} + b'_{0}.
\end{equation}
Here, \( \text{Branch}_x(u_1) \) and \( \text{Branch}_y(u_2) \) represent the feature vectors extracted from the boundary points for the \( x \)-coordinates and \( y \)-coordinates, respectively. The function \( \text{Trunk}(\xi, \eta) \) extracts high-dimensional features from the computational domain coordinates \( (\xi, \eta) \), shared by both the \( x \)-coordinate and \( y \)-coordinate generations. The bias terms \( b_{0} \) and \( b'_{0} \) correspond to the Branch-x and Branch-y sub-networks, adjusting the final values of the generated coordinates.
Through the inner product operation, information from the boundary conditions and the computational domain is fused to generate mesh coordinates \( (x, y) \) in the physical domain. 

\subsubsection{The Loss Function of MeshONet}
To ensure that the generated mesh satisfies the requirements, MeshONet employs a loss function based on interior and boundary points. The loss function in MeshONet consists primarily of two components: the interior loss term and the boundary loss term, which serve to ensure that the generated mesh is appropriately distributed within the physical domain and aligns with the boundary conditions.
The total loss function of MeshONet is defined as:  
\begin{equation}
L(\theta) = \alpha L_{\text{interior}} + \beta L_{\text{boundary}}  
\end{equation}  
where \( L_{\text{interior}} \) ensures that the distribution of generated mesh nodes within the physical domain aligns with the target mesh, and \( L_{\text{boundary}} \) enforces conformity with the input boundary conditions at the boundary points.  

The interior loss term is formulated as:  
\begin{equation}  
\begin{aligned}  
L_{\text{interior}} = &\sum_{j \in D'_{\text{interior}}} \left[ (\hat{G}_x(u_1,u_2)(\xi_j, \eta_j) - G_x(u_1,u_2)(\xi_j, \eta_j))^2 \right. \\
& + \left. (\hat{G}_y(u_1,u_2)(\xi_j, \eta_j) - G_y(u_1,u_2)(\xi_j, \eta_j))^2 \right]  
\end{aligned}  
\end{equation}

Similarly, the boundary loss term is defined as:  
\begin{equation}  
\begin{aligned}  
L_{\text{boundary}} = &\sum_{j \in D'_{\text{boundary}}} \left[ (\hat{G}_x(u_1,u_2)(\xi_j, \eta_j) - G_x(u_1,u_2)(\xi_j, \eta_j))^2 \right. \\
& + \left. (\hat{G}_y(u_1,u_2)(\xi_j, \eta_j) - G_y(u_1,u_2)(\xi_j, \eta_j))^2 \right]  
\end{aligned}  
\end{equation}

Here, \( D'_{\text{interior}} \) and \( D'_{\text{boundary}} \) denote the sets of sampled points in the interior and boundary regions of the computational domain, respectively, while \( \hat{G}_x \) and \( \hat{G}_y \) represent the approximated operators for generating the \( x \)- and \( y \)-coordinates.  Meanwhile, $\alpha$ and $\beta$ are weighting coefficients that balance the influence of the interior loss and boundary loss within the total loss. By minimizing this loss function, the model can approximate the nonlinear mapping relationship from the boundary conditions to the mesh nodes, thereby ensuring that the generated mesh satisfies the expected geometric and boundary requirements both within the physical domain and at the boundaries.
\begin{figure*}
	\centering
		\includegraphics[scale=0.515]{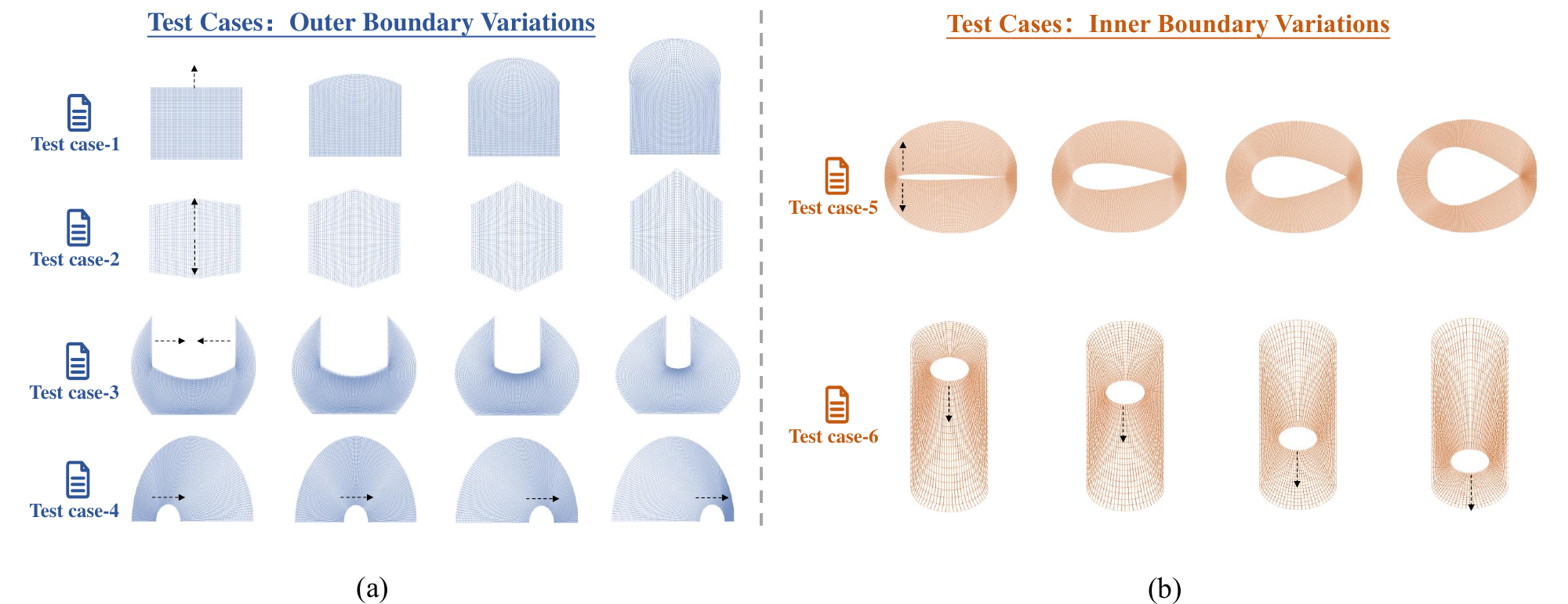}
	\caption{The figure presents the test cases, where (a) illustrates the test cases involving variations in the outer boundary, and (b) depicts the test cases with modifications to the inner boundary. The arrows indicate the direction of the corresponding changes.
}
	\label{FIG::dataset}
\end{figure*}

\section{Results}
In this section, we first describe the experimental setup and the test cases used to evaluate our method. We then provide a comparative analysis of our approach against modified operator learning-based methods to validate the feasibility of our method. Subsequently, we assess the model’s generalization ability across diverse geometries, as well as its performance in mesh refinement.
\subsection{Experiment Settings}
This experiment was conducted on a server equipped with an NVIDIA RTX 4090 GPU, using the PyTorch framework for model training and testing. To facilitate the reproduction of this work, we have provided the model's parameter settings in the appendix. In order to  compare the mesh quality generated by our method, we use TFI and PDE-based approaches as baseline benchmarks. These methods are well-established and widely used in the field, making them suitable for this evaluation.

Furthermore, as previously noted, operator learning-based methods  cannot directly apply to mesh generation tasks. To bridge this gap, we have made architectural modifications to several advanced operator learning-based methods, including DeepONet \cite{lu2021learning}, a deep learning-based operator learning framework designed for mapping input-output functions; POD-DeepONet \cite{lu2022comprehensive}, which combines proper orthogonal decomposition with DeepONet for enhanced efficiency and accuracy; FNO1D \cite{li2020fourier}, the Fourier neural operator designed for solving differential equations using Fourier transforms; and FNO2D \cite{li2020fourier}, an extension of FNO1D to handle 2D data.

\subsection{Test Cases}
To evaluate the generalization performance of MeshONet under diverse geometric conditions, we designed two types of test cases. The first type involves modifications to the outer boundary, including geometric variations such as changes in curvature, angle size, opening size, and the movement of the lower boundary semicircle. The second type focuses on alterations to the inner boundary, such as variations in the thickness of an airfoil or the movement of internal circular holes. 

As shown in Figure \ref{FIG::dataset}(a), Test Case-1 consists of arch shapes created by varying the curvature of the top boundary. Test Case-2 features hexagonal shapes, with modifications to the angles of both the top and bottom boundaries. Test Case-3 includes wrenches with different upper boundary opening sizes, while Test Case-4 contains shapes where the lower boundary semicircle is progressively shifted to the right. In addition to assessing the model’s generalization across variations in the outer boundary, we also evaluated its performance with respect to changes in the inner boundary. Test Case-5 consists of airfoils with varying thicknesses, generated by modifying the internal boundary. Test Case-6 involves shapes with internal circular holes, where the position of the holes is varied through vertical displacements.
\subsection{Comparison of Modified Operator Learning-Based methods}
 To validate the feasibility of the proposed model architecture, we conducted a comparative analysis with several operator learning-based approaches, including DeepONet \cite{lu2021learning}, POD-DeepONet \cite{lu2022comprehensive}, FNO1D \cite{li2020fourier}, and FNO2D \cite{li2020fourier}. Given that traditional operator methods are incapable of generating meshes, we implemented architectural modifications to these approaches. Further details of the experimental setup, network architecture modifications, and specific experimental results are provided in the appendix.

As shown in Figure \ref{FIG:loss}, MeshONet consistently maintains the lowest loss, demonstrating a significant difference in magnitude compared to other operator learning-based methods. This result highlights the effectiveness of our architecture in addressing the multivariable mapping problem, specifically in the context of mesh generation.

\begin{figure}[t]
	\centering
		\includegraphics[width = 0.49\textwidth]{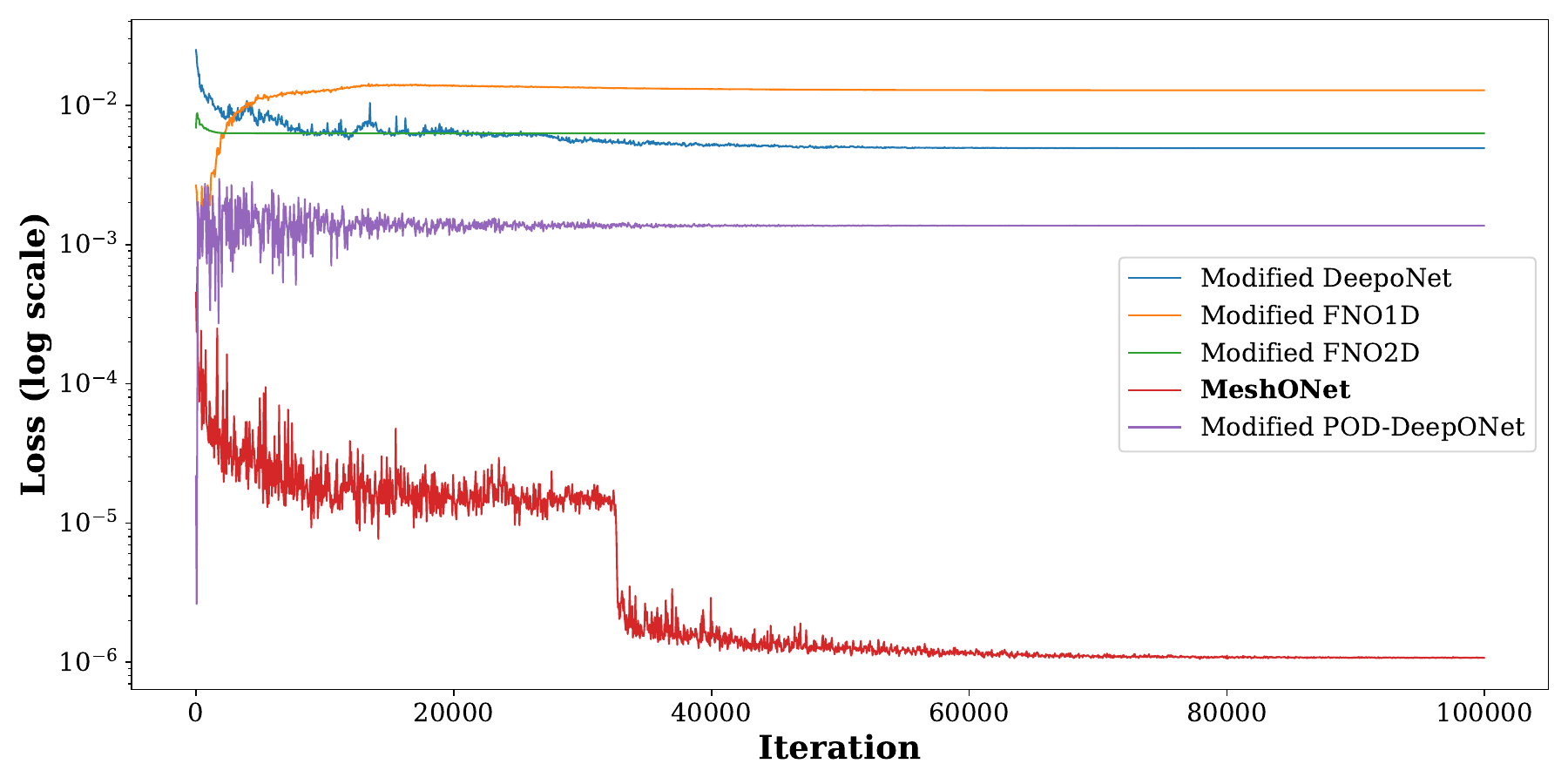}
	\caption{Loss comparison across different operator learning-based methods over 100,000 iterations, displayed on a logarithmic scale.}
	\label{FIG:loss}
\end{figure}
\subsection{Experiments Based on Outer Boundary Variations}
In this section, we  evaluate the model's generalization capability with respect to variations in the outer boundary conditions, conducting experiments on four distinct test cases, each corresponding to different outer boundary modifications.

\subsubsection{Experimental Results on Test Case-1}
In test case-1, we generated a series of arch samples by  adjusting the curvature of the top boundary. To evaluate the model's generalization capability, we conducted two types of experiments: interpolation and extrapolation. In the interpolation experiment, the model was trained on samples with the minimal and maximal curvatures and tested on samples with intermediate curvatures. This setup allows the model to learn from extreme configurations and tests its performance within an unobserved range of curvatures. The extrapolation experiment, being more challenging, involved training the model on samples with low curvature and testing it on samples with high curvature, thereby assessing its ability to generalize beyond the training range.

\begin{figure}[t]
    \centering
    % 第一张图
    \begin{minipage}[b]{0.49\textwidth}
        \centering
        \includegraphics[width=\textwidth, height=5.2cm]{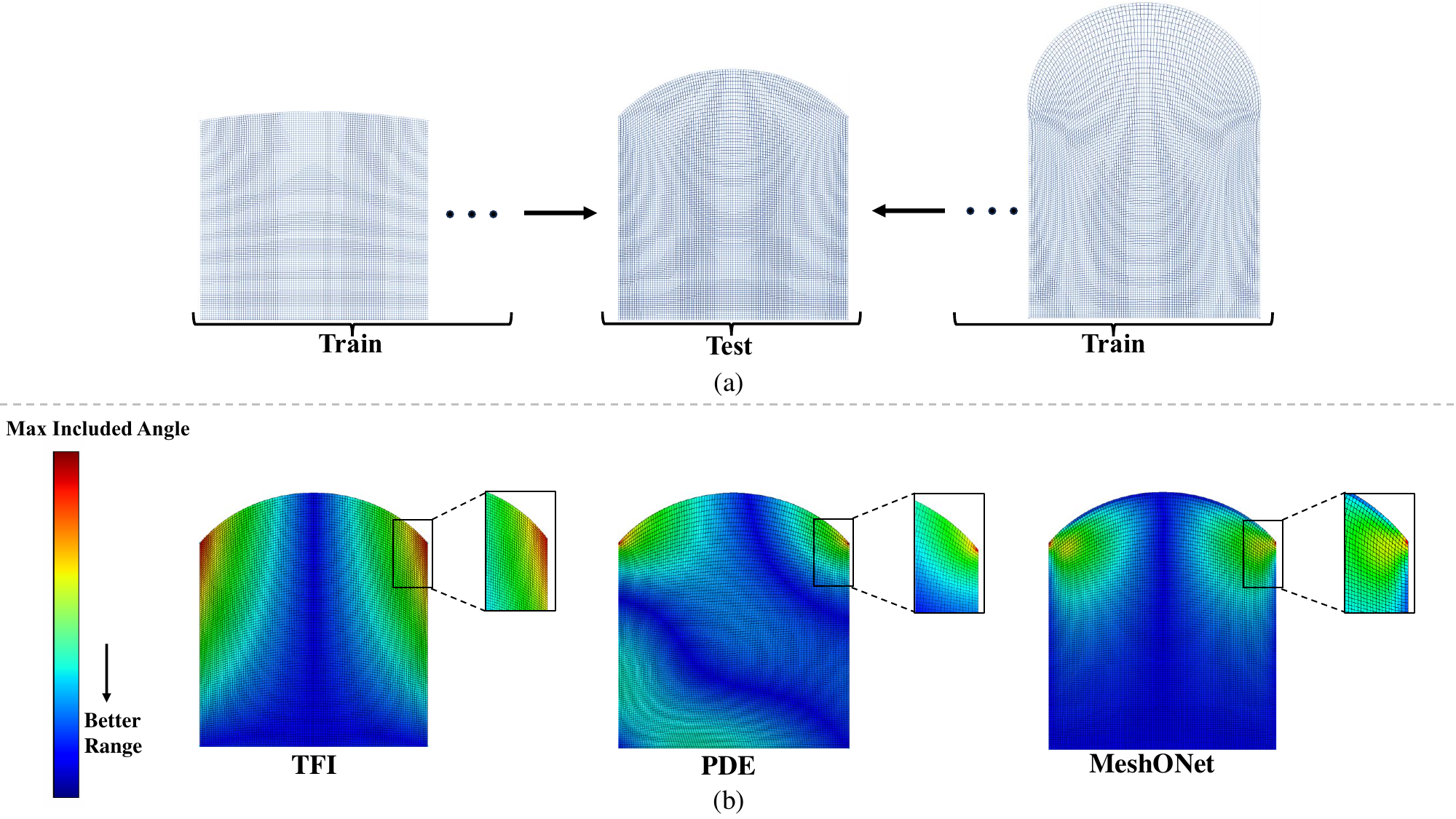}
        \caption{Interpolation experiment on Test Case-1. (a) shows the setup, with training on samples of low and high curvature, and testing on intermediate curvature. (b) compares mesh quality between MeshONet, TFI, and PDE methods, with the color map showing the maximum included angle, where blue represents higher quality.}
        \label{fig:in-gx}
    \end{minipage}

    % 第二张图
    \begin{minipage}[b]{0.49\textwidth}
        \centering
        \includegraphics[width=\textwidth, height=5.2cm]{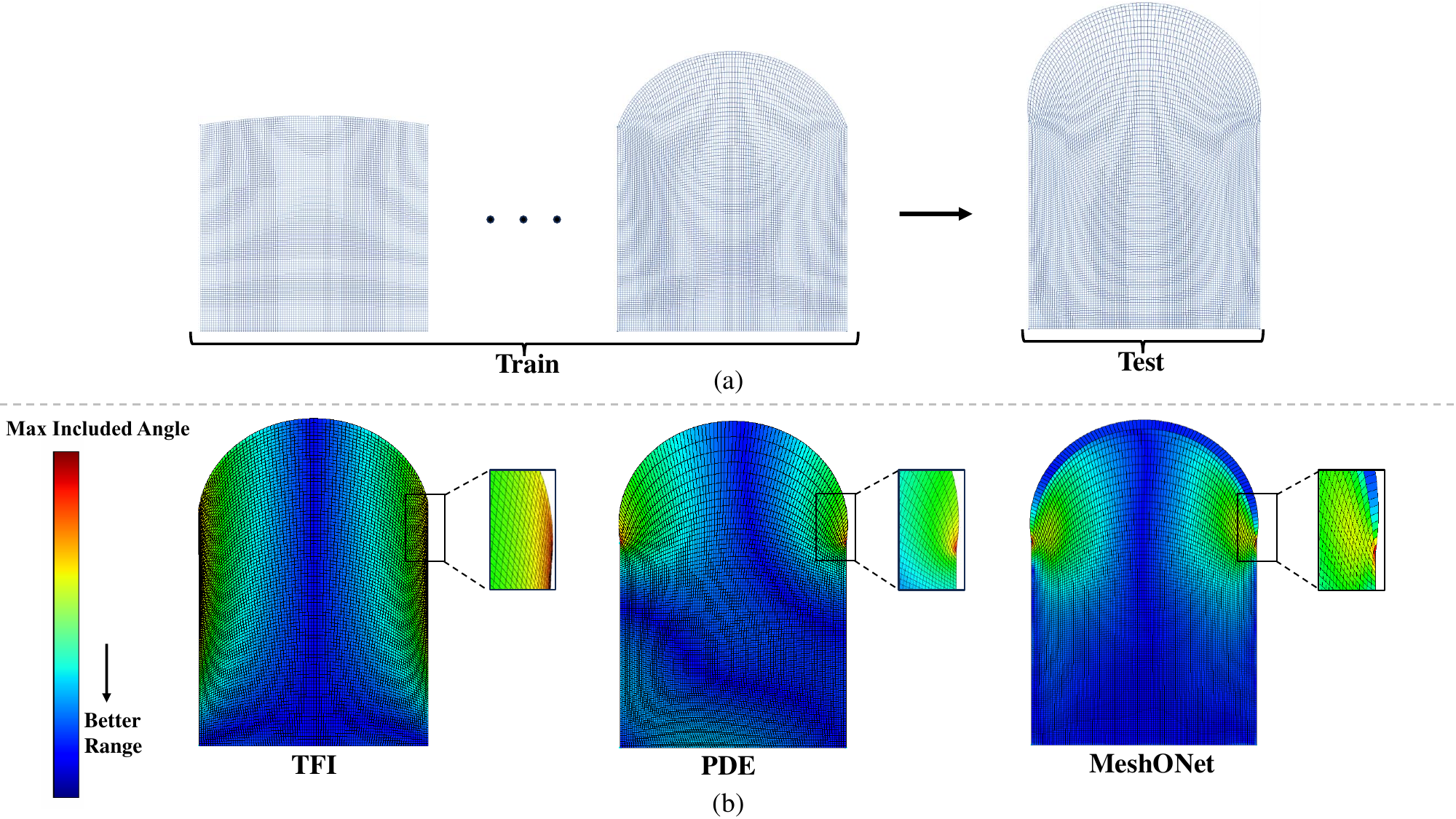}
        \caption{Extrapolation experiment on Test Case-1. (a) shows the setup, with training on lower curvature samples and testing on higher curvature samples. (b) compares mesh quality between MeshONet, TFI, and PDE methods, with the color map indicating the maximum included angle, where blue represents higher quality.}
        \label{fig:ex-gx}
    \end{minipage}  
    \label{fig:combined-gx}
\end{figure}

As shown in Figures \ref{fig:in-gx}(b) and \ref{fig:ex-gx}(b), the model exhibits outstanding performance in both interpolation and extrapolation scenarios, achieving mesh quality that exceeds that of traditional methods. Additionally, it is worth noting that the mesh quality around the corner points of the arch test case tends to be relatively poor. However, by zooming in on the mesh at these corner regions, we observe that our method performs exceptionally well. 

\subsubsection{Experimental Results on Test Case-2}

\begin{figure}[t]
    \centering
    % 第一张图
    \begin{minipage}[b]{0.49\textwidth}
        \centering
        \includegraphics[width=\textwidth, height=5.2cm]{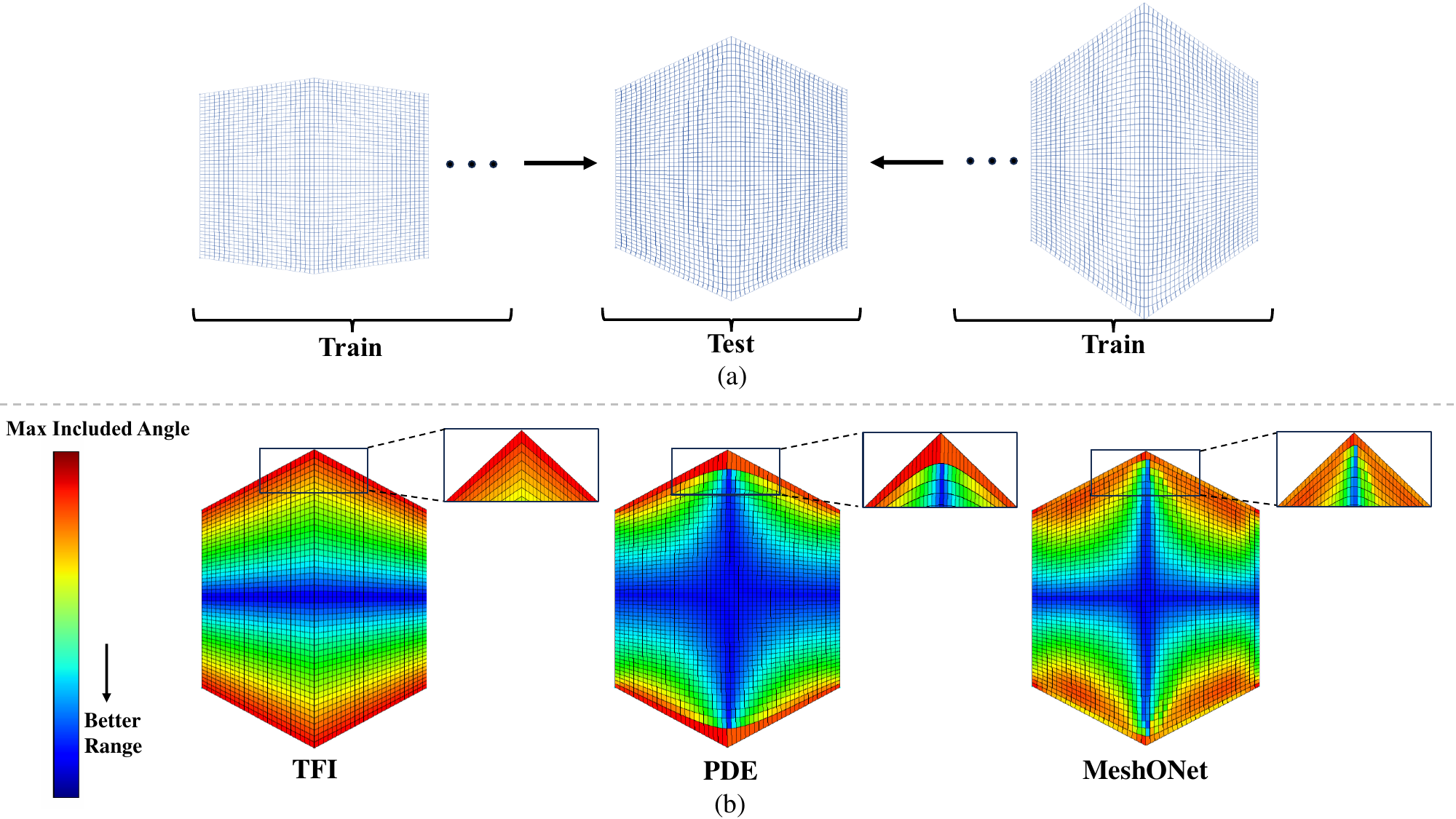}
        \caption{Interpolation experiment on Test Case-2. (a) shows the setup, with training on samples with large and small angle sizes on the top and bottom boundaries, and testing on intermediate angle sizes. (b) compares mesh quality between MeshONet, TFI, and PDE methods, with the color map indicating the maximum included angle, where blue represents higher quality.}
        \label{fig:in-lx}
    \end{minipage}

    % 第二张图
    \begin{minipage}[b]{0.49\textwidth}
        \centering
        \includegraphics[width=\textwidth, height=5.2cm]{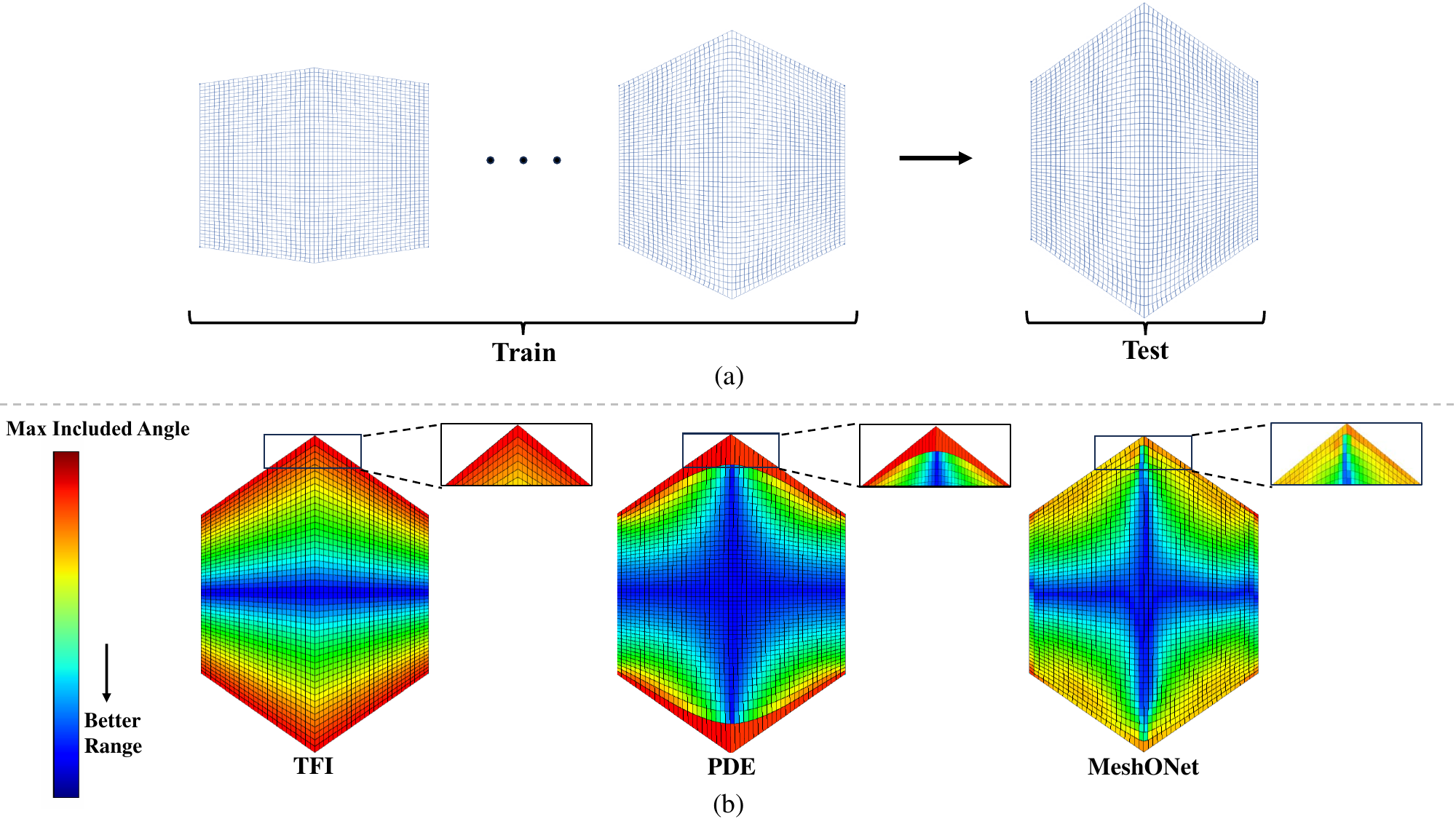}
        \caption{Extrapolation experiment on Test Case-2. (a) shows the setup, with training on samples with smaller angles on the top and bottom boundaries, and testing on samples with larger angles to evaluate extrapolation. (b) compares mesh quality between MeshONet, TFI, and PDE methods, with the color map indicating the maximum included angle, where blue represents higher quality.}
        \label{fig:ex-lx}
    \end{minipage}
    \label{fig:combined-lx}
\end{figure}

In test case-2, we used hexagonal samples with various shapes by simultaneously adjusting the angles of the top and bottom boundaries. As shown in Figures \ref{fig:in-lx}(a) and \ref{fig:ex-lx}(a), in the interpolation experiment, we selected samples with large and small angles of the top and bottom boundaries for training and used samples with intermediate angles of the top and bottom boundaries for testing. In the extrapolation experiment, we selected samples with larger angles of the top and bottom boundaries for training and samples with smaller angles of the top and bottom boundaries for testing.

As shown in Figures \ref{fig:in-lx}(b) and \ref{fig:ex-lx}(b), the model performs exceptionally well on the test samples, accurately capturing the geometric features of hexagonal shapes during variations in the angles between the top and bottom boundaries. By zooming in on the mesh at the top, it can be observed that the mesh generated by the PDE method lacks smoothness on this shape, while the mesh produced by TFI exhibits overall poor quality. In contrast, our method excels in this test case, demonstrating exceptional performance in both interpolation and extrapolation, with outstanding generalization across a range of variations in geometric angles. 

\begin{figure}[t]
    \centering
    % 第一张图
    \begin{minipage}[b]{0.49\textwidth}
        \centering
        \includegraphics[width=\textwidth, height=5.2cm]{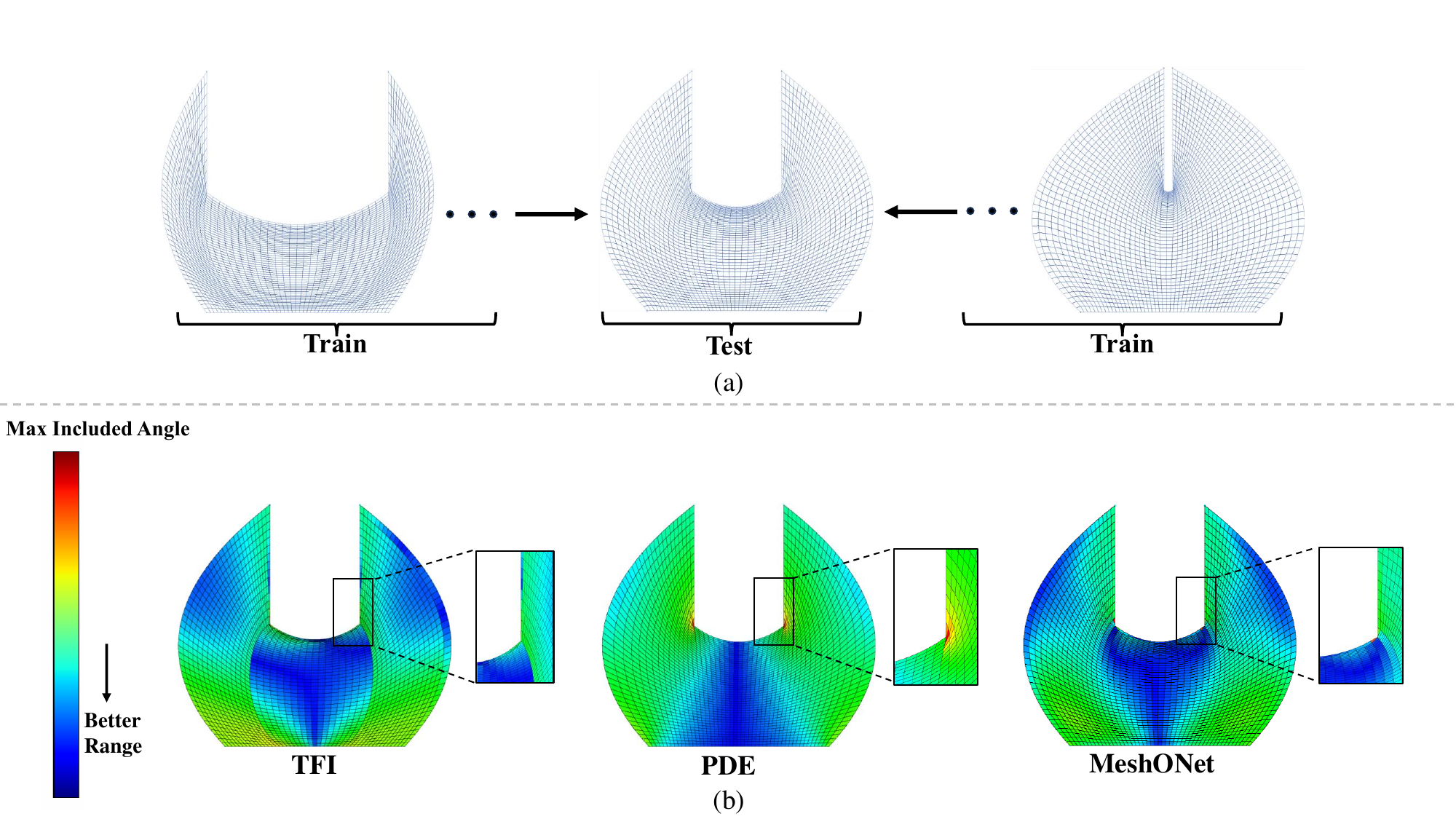}
        \caption{Interpolation experiment on Test Case-3. (a) shows the setup, with training on samples with large and small opening sizes, and testing on samples with intermediate sizes. (b) compares mesh quality between MeshONet, TFI, and PDE methods, with the color map indicating the maximum included angle, where blue represents higher quality.}
        \label{fig:in-bs}
    \end{minipage}

    % 第二张图
    \begin{minipage}[b]{0.49\textwidth}
        \centering
        \includegraphics[width=\textwidth, height=5.2cm]{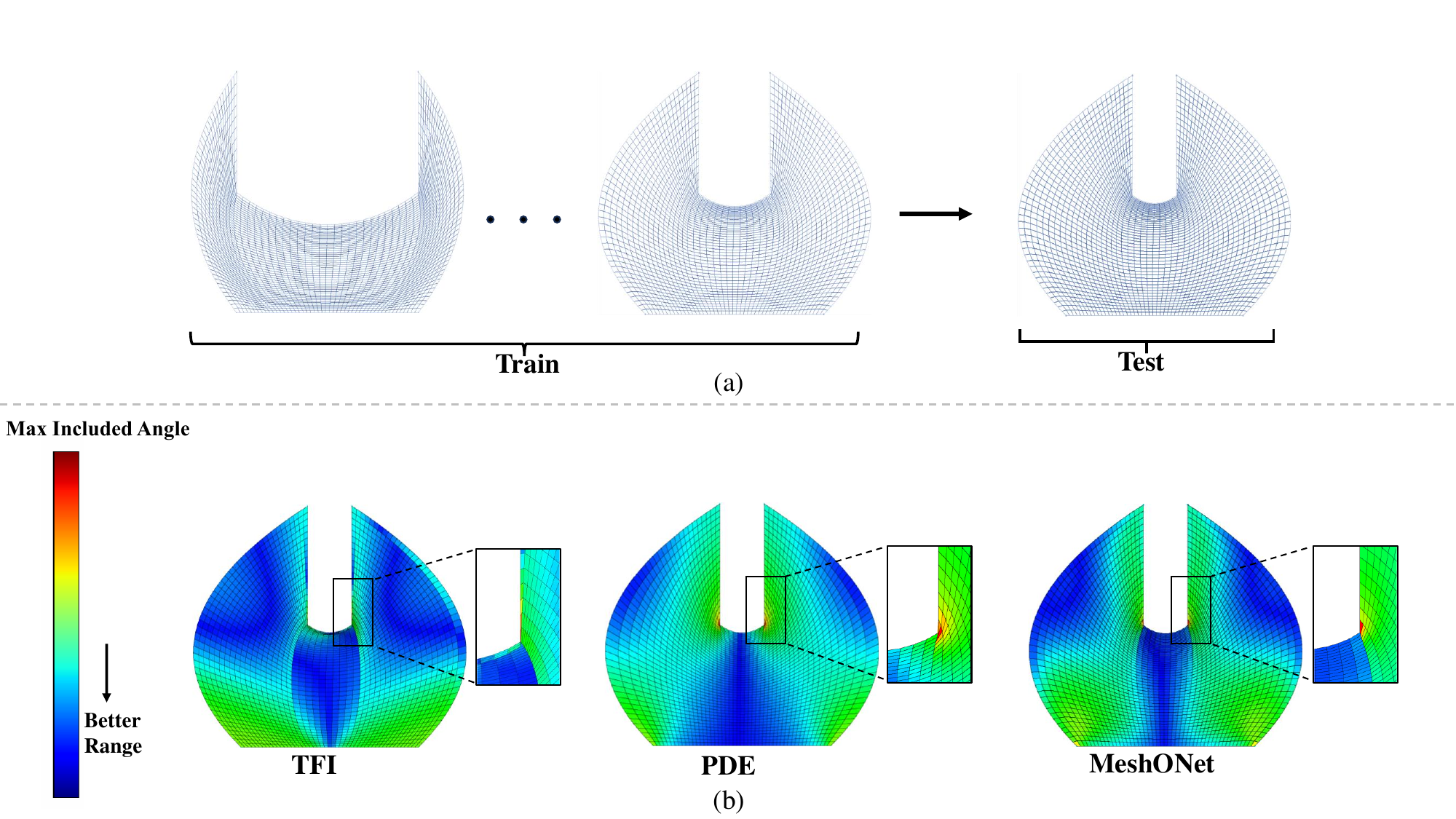}
        \caption{Extrapolation experiment on Test Case-3. (a) shows the setup, with training on samples with large opening sizes and testing on smaller ones. (b) compares mesh quality between MeshONet, TFI, and PDE methods, with the color map indicating the maximum included angle, where blue represents higher quality.}
        \label{fig:ex-bs}
    \end{minipage}
    \label{fig:combined-bs}
\end{figure}

\subsubsection{Experimental Results on Test Case-3}
In test case-3, a series of wrenches were generated by varying the opening size of the top boundary to simulate practical engineering requirements. Two types of experiments were then conducted on these samples: interpolation experiments and extrapolation experiments, to evaluate the model's generalization capability on this test case. As illustrated in Figures \ref{fig:in-bs}(a) and \ref{fig:ex-bs}(a), in the interpolation experiment, samples with larger and smaller opening sizes were used for training, while samples with intermediate opening sizes served as the testing set. In the extrapolation experiment, training was performed on samples with larger opening sizes, and testing was conducted using samples with smaller opening sizes.

As shown in Figures \ref{fig:in-bs}(b) and \ref{fig:ex-bs}(b), MeshONet generates meshes comparable to traditional methods. In interpolation, it matches TFI and outperforms PDE methods. In extrapolation, it slightly trails TFI but still outperforms PDE. MeshONet's ability to maintain high mesh quality across varied conditions demonstrates its potential for real-world applications with frequent geometric variations.

\subsubsection{Experimental Results on Test Case-4}
In test case-4, we consider the scenario where the horizontal position of the lower semicircular boundary varies, leading to corresponding geometric changes. A series of samples are generated by  altering the horizontal position of the lower semicircle from left to right. For the experiments, we randomly select samples from various positions for training, while reserving the remaining positional samples for testing. The goal is to assess the model's ability to generalize mesh generation across different horizontal positions. This experimental setup enables a comprehensive evaluation of the model's performance in handling positional variations.

\begin{figure}[t]
	\centering
 		\includegraphics[width =  0.49\textwidth,height=5cm]{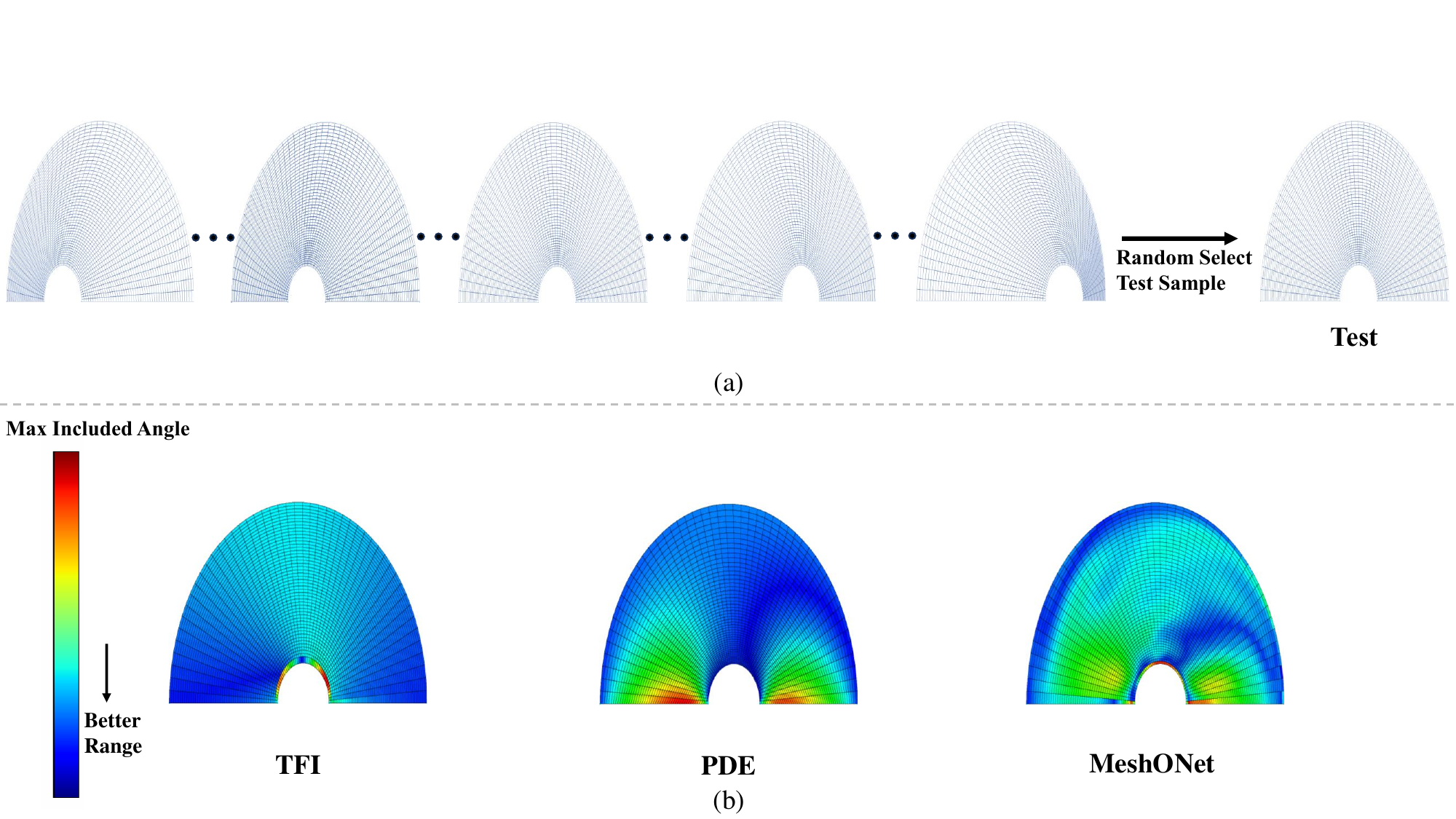}
	\caption{The position adjustment experiment on Test Case-4. (a) indicates that one sample is randomly selected for testing, while the remaining samples are used for training. (b) shows a comparison of mesh quality between meshes generated by MeshONet and those generated by TFI and PDE methods. The color map indicates the maximum included angle, with blue representing higher mesh quality.}
	\label{FIG:改变半圆位置}
\end{figure}

As shown in Figure \ref{FIG:改变半圆位置}, our model exhibits robust generalization capabilities, even when confronted with geometric variations induced by the horizontal displacement of the lower semicircular boundary. Although its performance may not consistently surpass that of traditional methods in some cases, the generated mesh quality remains highly satisfactory.

\subsection{Experiments Based on Inner Boundary Variations}
In this section, we  evaluate the model's generalization capability with respect to variations in the inner boundary conditions, conducting experiments on two distinct test cases representing different geometric configurations.

\subsubsection{Experimental Results on Test Case-5}
\begin{figure}[t]
    \centering
    % 第一张图
    \begin{minipage}[b]{0.45\textwidth}
        \centering
        \includegraphics[width=\textwidth, height=5cm]{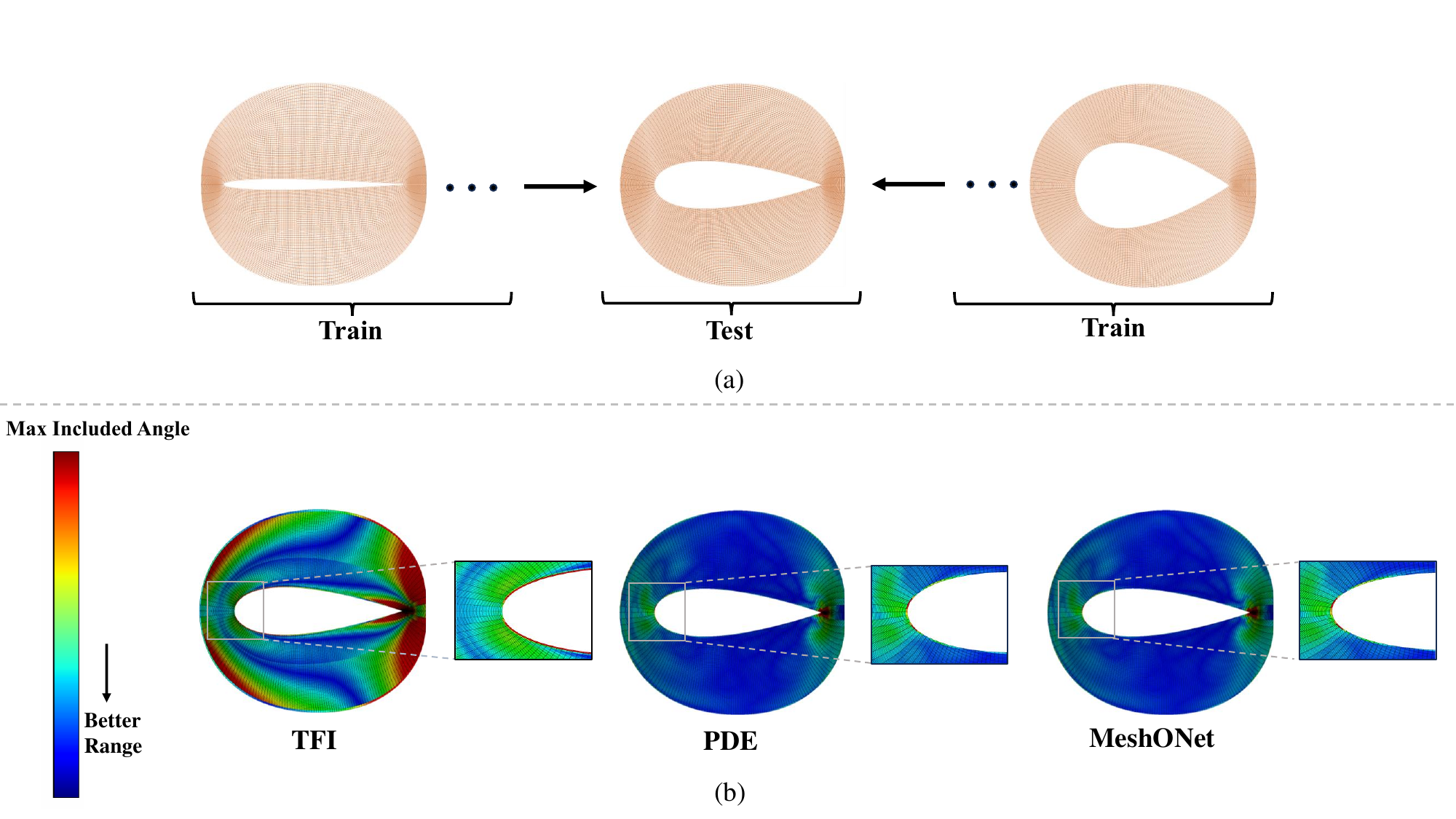}
        \caption{Interpolation experiment on Test Case-5. (a) shows the setup, with training on thin and thick airfoil shapes, and testing on samples with intermediate thickness. (b) compares mesh quality between MeshONet, TFI, and PDE methods, with the color map indicating the maximum included angle, where blue represents higher quality.}
        \label{fig:in-yx}
    \end{minipage}
    % 第二张图
    \begin{minipage}[b]{0.45\textwidth}
        \centering
        \includegraphics[width=\textwidth, height=5cm]{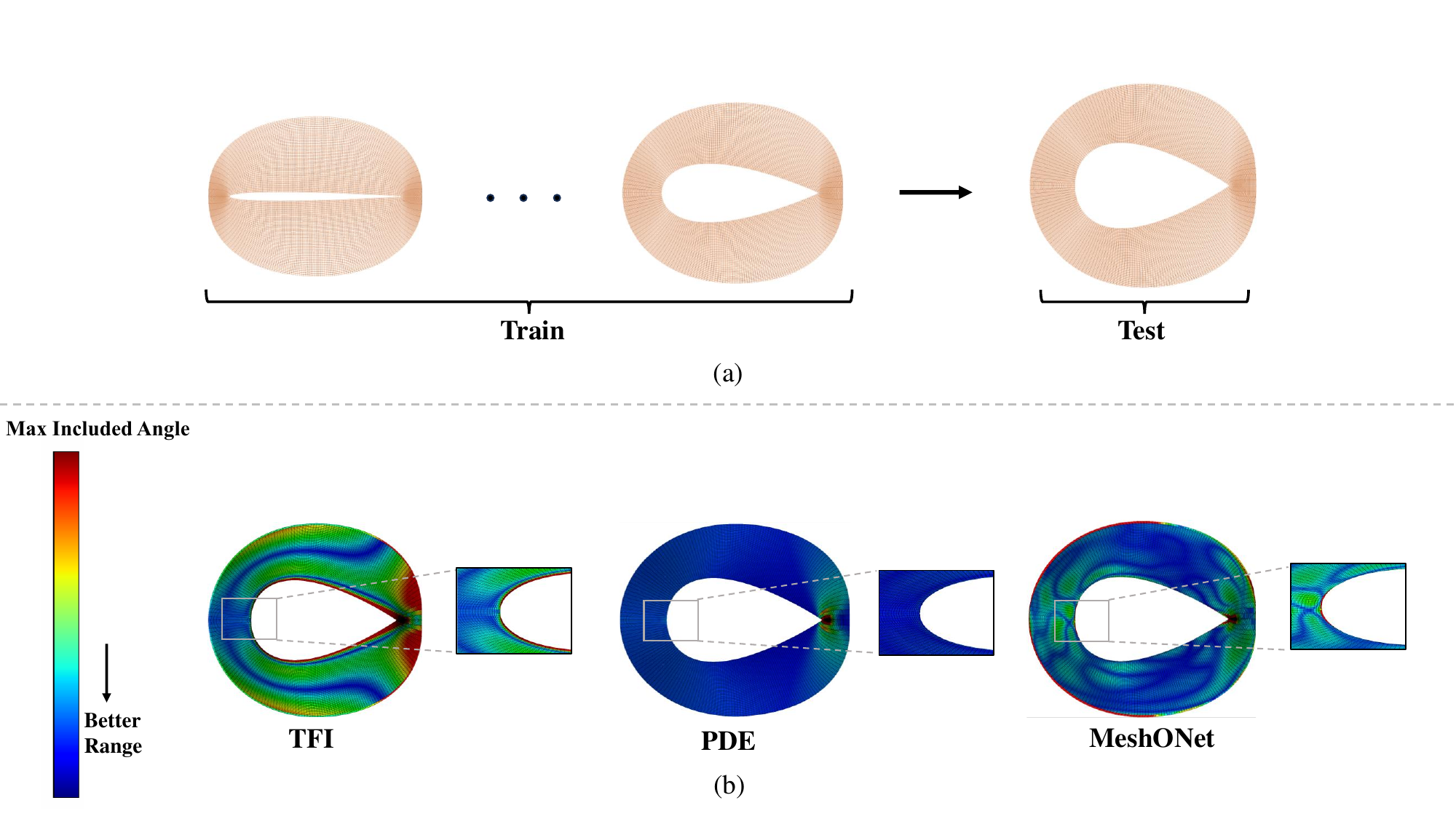}
        \caption{Extrapolation experiment on Test Case-5. (a) shows the setup, with training on thinner airfoils and testing on thicker ones to evaluate extrapolation. (b) compares mesh quality between MeshONet, TFI, and PDE methods, with the color map indicating the maximum included angle, where blue represents higher quality.}
        \label{fig:ex-yx}
    \end{minipage}
    \label{fig:combined-yx}
\end{figure}

In test case-5, we generated a series of airfoil samples with varying thicknesses, where the thickness change corresponds to modifications in the inner boundary. To evaluate the model's generalization ability across different geometries, we conducted both interpolation and extrapolation experiments. As shown in Figures \ref{fig:in-yx}(a) and \ref{fig:ex-yx}(a), in the interpolation experiment, we used samples with smaller and larger thicknesses for training, and samples with intermediate thickness for testing. In the extrapolation experiment, we used samples with smaller thicknesses for training and samples with larger thicknesses for testing.

As shown in Figures \ref{fig:in-yx}(b) and \ref{fig:ex-yx}(b), in the interpolation task, our method matches the PDE method in mesh quality, outperforming TFI. In the extrapolation task, it slightly lags behind the PDE method but still outperforms TFI, demonstrating the robustness of our model for both tasks, with strong potential for practical engineering applications.

\subsubsection{Experimental Results on Test Case-6}
In test case-6, we consider the scenario where the variation in the position of the internal circular hole leads to geometric changes. A series of samples are generated by  altering the vertical position of the hole from top to bottom. For the experiments, we randomly select samples from various vertical positions for training, while reserving the remaining positional samples for testing. The goal is to assess the model's ability to generalize mesh generation across different vertical positions.

\begin{figure}[t]
	\centering
 		\includegraphics[width =  0.49\textwidth,height=4.8cm]{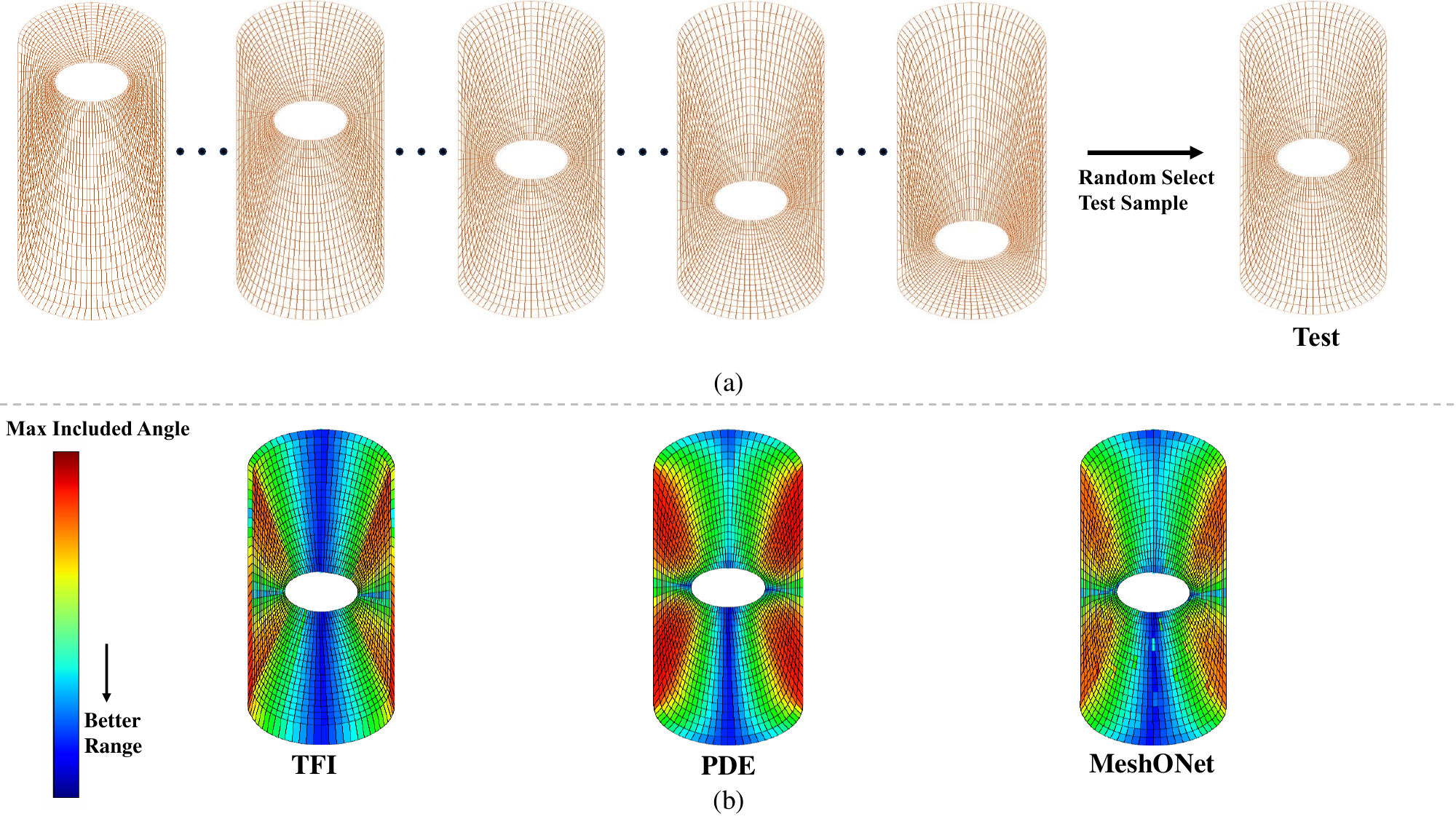}
	\caption{Inner boundary adjustment for Test Case-6: (a) shows one sample selected for testing and the rest for training. (b) compares the mesh quality generated by MeshONet, TFI, and PDE methods, with the color map representing the maximum included angle, where blue indicates higher quality.}
	\label{FIG:开瓶器}
\end{figure}

As shown in Figure \ref{FIG:开瓶器}, our model performs exceptionally well in scenarios where geometry is altered through vertical displacement, with mesh quality surpassing that of traditional methods.

\subsection{Mesh Refinement Experiment}
\begin{figure}[t]
	\centering
		\includegraphics[width = 0.49\textwidth,height=5cm]{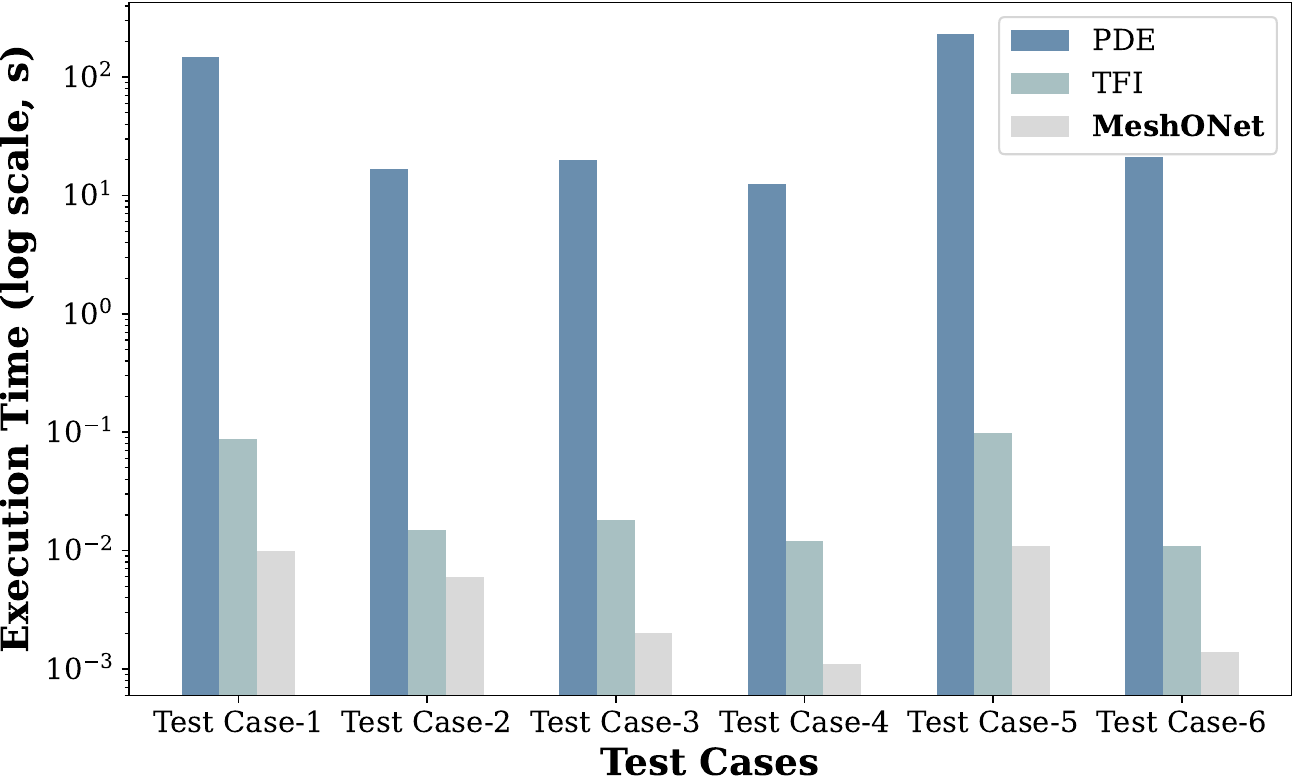}
	\caption{Execution time comparison for mesh generation across different test cases and methods, shown on a log scale. }
	\label{FIG:time_in_ex}
\end{figure}

Figure \ref{FIG:time_in_ex} compares the execution time for mesh generation across different test cases and methods. As shown, our method outperforms traditional methods, achieving up to a four-order-of-magnitude improvement in efficiency. We also evaluate the model's performance in mesh refinement, focusing on its ability to refine meshes while maintaining high quality and improving refinement efficiency. Refining the mesh enables the capture of finer geometric details and boundary features, which reduces numerical errors and enhances the accuracy of simulations.

\begin{figure}[t]
	\centering
		\includegraphics[width = 0.49\textwidth,height=5.2cm]{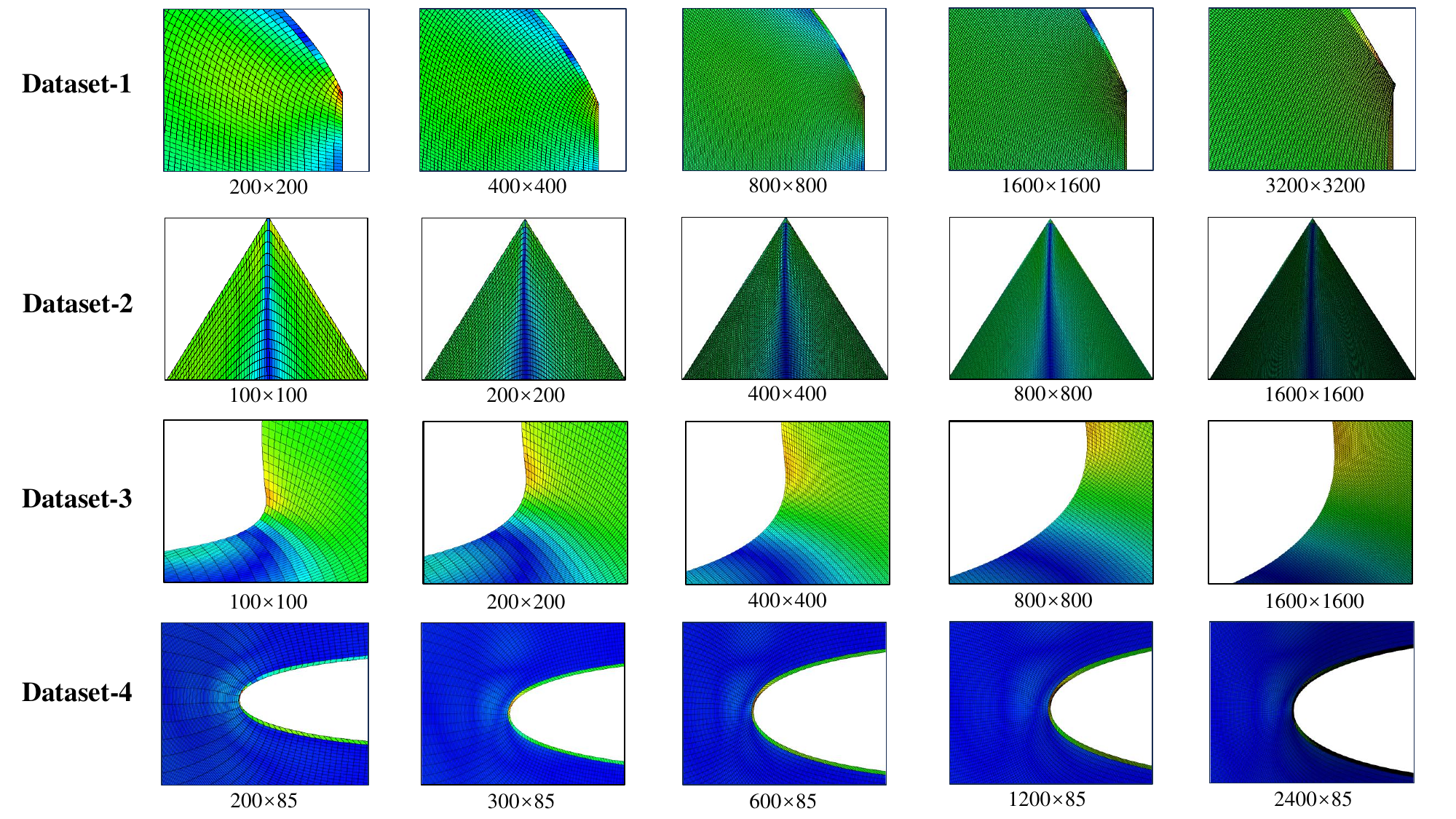}
	\caption{Mesh refinement experiment on four test cases with increasing resolutions.}
	\label{FIG:super}
\end{figure}

The experimental results, as shown in Table \ref{tab::super} and Figure \ref{FIG:super}, further validate the capabilities of our model. Specifically, Table \ref{tab::super} presents the generation time across varying mesh densities, revealing that the time required for both the TFI method and the PDE method increases rapidly as the resolution increases. In contrast, our method shows little variation in execution time, even as the resolution increases. Notably, the PDE method fails to generate a mesh within the specified time frame at resolutions of 1600x1600 and 3200x3200. This highlights the superiority of our approach in handling large-scale meshes. Furthermore, as shown in Figure \ref{FIG:super}, the high-resolution meshes generated by our model confirm its  effectiveness in mesh refinement tasks. It can be observed that, even as the mesh resolution increases, the quality of the meshes generated by our model remains consistently high, further demonstrating its robustness in handling large-scale mesh generation tasks.
\begin{table}[b]
    \centering
    \renewcommand{\arraystretch}{1.5} % 调整表格行高
    \caption{Meshing overhead comparison for evaluating mesh refinement capabilities. The `-' symbol indicates excessively long meshing overhead.}
    \begin{tabular}{c @{\hskip 5pt} c @{\hskip 5pt} c @{\hskip 5pt} c @{\hskip 5pt} c @{\hskip 5pt} c @{\hskip 5pt} c}
        \hline
       Mesh Size & $100^2$ & $200^2$ & $400^2$ & $800^2$ & $1600^2$ & $3200^2$ \\
        \hline
        TFI      & 0.087s  & 0.218s  & 0.928s  & 3.817s  & 15.238s  & 72.121s \\
        PDE      & 140.331s & 586.170s & 2360.705s & 9795.884s  & - & - \\
        \textbf{MeshONet} & \textbf{0.001s} & \textbf{0.002s} & \textbf{0.002s} & \textbf{0.006s} & \textbf{0.026s} & \textbf{0.278s}  \\
        \hline
    \label{tab::super}
    \end{tabular}
\end{table}

\vspace{2cm}
\section{Conclusion}
This paper introduces MeshONet, a generalizable and efficient method for structured mesh generation based on operator learning. The method transforms the mesh generation task into an operator learning problem with multivariable mapping, involving multiple input and solution functions. By developing a specialized operator learning architecture for mesh generation, which incorporates a dual-branch structure with a shared trunk, MeshONet effectively learns operators that capture the underlying meshing rules. Additionally, it addresses the inherent challenges of multivariable mapping, providing a framework that can serve as a reference for other applications involving multi-input, multi-output tasks.

The experimental results show that MeshONet successfully addresses the challenges of balancing efficiency and quality in traditional methods, achieving up to a four-order-of-magnitude improvement in mesh generation efficiency while maintaining high mesh quality. It also overcomes the limited generalization capabilities of physics-informed methods, enabling generalization across geometric variations and eliminating the need for retraining. Moreover, MeshONet performs exceptionally well in mesh refinement tasks, effectively training on low-resolution samples and generating high-quality meshes for high-resolution evaluations, with minimal impact on inference time.

Despite these advancements, some limitations remain. Specifically, the model's parameter size is determined by the number of boundary points, which in turn constrains its generalization capacity. An excessive number of boundary points may lead to overfitting, and too few boundary points may result in underfitting. Therefore, extending the model to 3D problems will introduce additional challenges. Future work will focus on refining the model architecture to mitigate the impact of boundary sampling points on parameter size, thereby enhancing both efficiency and generalization performance, and will also extend the model to 3D structured mesh generation.

\bibliographystyle{IEEEtran}
\bibliography{main}

\begin{thebibliography}{10}
\providecommand{\url}[1]{#1}
\csname url@samestyle\endcsname
\providecommand{\newblock}{\relax}
\providecommand{\bibinfo}[2]{#2}
\providecommand{\BIBentrySTDinterwordspacing}{\spaceskip=0pt\relax}
\providecommand{\BIBentryALTinterwordstretchfactor}{4}
\providecommand{\BIBentryALTinterwordspacing}{\spaceskip=\fontdimen2\font plus
\BIBentryALTinterwordstretchfactor\fontdimen3\font minus \fontdimen4\font\relax}
\providecommand{\BIBforeignlanguage}[2]{{%
\expandafter\ifx\csname l@#1\endcsname\relax
\typeout{** WARNING: IEEEtran.bst: No hyphenation pattern has been}%
\typeout{** loaded for the language `#1'. Using the pattern for}%
\typeout{** the default language instead.}%
\else
\language=\csname l@#1\endcsname
\fi
#2}}
\providecommand{\BIBdecl}{\relax}
\BIBdecl

\bibitem{blockley2010encyclopedia}
R.~Blockley and W.~Shyy, \emph{Encyclopedia of aerospace engineering}.\hskip 1em plus 0.5em minus 0.4em\relax American Institute of Aeronautics and Astronautics, Inc., 2010.

\bibitem{crolla2015encyclopedia}
D.~Crolla, D.~G. Foster, T.~Kobayashi, and N.~Vaughan, \emph{Encyclopedia of automotive engineering}.\hskip 1em plus 0.5em minus 0.4em\relax John Wiley \& Sons, 2015.

\bibitem{petterssen2011introduction}
S.~Petterssen, \emph{Introduction to meteorology}.\hskip 1em plus 0.5em minus 0.4em\relax Read Books Ltd, 2011.

\bibitem{mccormick2009ocean}
M.~E. McCormick, \emph{Ocean engineering mechanics: with applications}.\hskip 1em plus 0.5em minus 0.4em\relax Cambridge University Press, 2009.

\bibitem{callister2020materials}
W.~D. Callister~Jr and D.~G. Rethwisch, \emph{Materials science and engineering: an introduction}.\hskip 1em plus 0.5em minus 0.4em\relax John wiley \& sons, 2020.

\bibitem{rvachev2001transfinite}
V.~L. Rvachev, T.~I. Sheiko, V.~Shapiro, and I.~Tsukanov, ``Transfinite interpolation over implicitly defined sets,'' \emph{Computer aided geometric design}, vol.~18, no.~3, pp. 195--220, 2001.

\bibitem{babuska2012modeling}
I.~Babuska, J.~E. Flaherty, W.~D. Henshaw, J.~E. Hopcroft, J.~E. Oliger, and T.~Tezduyar, \emph{Modeling, mesh generation, and adaptive numerical methods for partial differential equations}.\hskip 1em plus 0.5em minus 0.4em\relax Springer Science \& Business Media, 2012, vol.~75.

\bibitem{chen2022mgnet}
X.~Chen, T.~Li, Q.~Wan, X.~He, C.~Gong, Y.~Pang, and J.~Liu, ``{MGNet}: a novel differential mesh generation method based on unsupervised neural networks,'' \emph{Engineering with Computers}, vol.~38, no.~5, pp. 4409--4421, 2022.

\bibitem{peng20243dmeshnet}
J.~Peng, X.~Chen, and J.~Liu, ``{3DMeshNet}: A three-dimensional differential neural network for structured mesh generation,'' \emph{arXiv preprint arXiv:2407.01560}, 2024.

\bibitem{effros2022theory}
E.~G. Effros and Z.-J. Ruan, \emph{Theory of Operator Spaces}.\hskip 1em plus 0.5em minus 0.4em\relax American Mathematical Society, 2022, vol. 386.

\bibitem{nguyen1989applications}
H.~L. Nguyen, ``On the applications of algebraic grid generation methods based on transfinite interpolation,'' Tech. Rep., 1989.

\bibitem{moin1998direct}
P.~Moin and K.~Mahesh, ``Direct numerical simulation: a tool in turbulence research,'' \emph{Annual review of fluid mechanics}, vol.~30, no.~1, pp. 539--578, 1998.

\bibitem{kaimal1994atmospheric}
J.~C. Kaimal and J.~J. Finnigan, \emph{Atmospheric boundary layer flows: their structure and measurement}.\hskip 1em plus 0.5em minus 0.4em\relax Oxford university press, 1994.

\bibitem{2002Automatic}
S.~Alfonzetti, S.~Coco, S.~Cavalieri, and M.~Malgeri, ``Automatic mesh generation by the let-it-grow neural network,'' \emph{IEEE Transactions on Magnetics}, vol.~32, no.~3, pp. 1349--1352, 2002.

\bibitem{ahmet2002neural}
{\c{C}}.~Ahmet and A.~Ahmet, ``Neural networks based mesh generation method in 2-d,'' in \emph{Eurasian Conference on Information and Communication Technology}.\hskip 1em plus 0.5em minus 0.4em\relax Springer, 2002, pp. 395--401.

\bibitem{lowther1993density}
D.~Lowther and D.~Dyck, ``A density driven mesh generator guided by a neural network,'' \emph{IEEE transactions on magnetics}, vol.~29, no.~2, pp. 1927--1930, 1993.

\bibitem{zhang2020meshingnet}
Z.~Zhang, Y.~Wang, P.~K. Jimack, and H.~Wang, ``{MeshingNet}: A new mesh generation method based on deep learning,'' in \emph{International conference on computational science}.\hskip 1em plus 0.5em minus 0.4em\relax Springer, 2020, pp. 186--198.

\bibitem{papagiannopoulos2021teach}
A.~Papagiannopoulos, P.~Clausen, and F.~Avellan, ``How to teach neural networks to mesh: Application on 2-d simplicial contours,'' \emph{Neural Networks}, vol. 136, pp. 152--179, 2021.

\bibitem{raissi2019physics}
M.~Raissi, P.~Perdikaris, and G.~E. Karniadakis, ``Physics-informed neural networks: A deep learning framework for solving forward and inverse problems involving nonlinear partial differential equations,'' \emph{Journal of Computational physics}, vol. 378, pp. 686--707, 2019.

\bibitem{kapoor2023physics}
T.~Kapoor, H.~Wang, A.~N{\'u}{\~n}ez, and R.~Dollevoet, ``Physics-informed neural networks for solving forward and inverse problems in complex beam systems,'' \emph{IEEE Transactions on Neural Networks and Learning Systems}, 2023.

\bibitem{hua2023physics}
J.~Hua, Y.~Li, C.~Liu, P.~Wan, and X.~Liu, ``Physics-informed neural networks with weighted losses by uncertainty evaluation for accurate and stable prediction of manufacturing systems,'' \emph{IEEE Transactions on Neural Networks and Learning Systems}, 2023.

\bibitem{chen2023developing}
X.~Chen, J.~Liu, Q.~Zhang, J.~Liu, Q.~Wang, L.~Deng, and Y.~Pang, ``Developing a novel structured mesh generation method based on deep neural networks,'' \emph{Physics of Fluids}, vol.~35, no.~9, 2023.

\bibitem{chen2022improved}
X.~Chen, J.~Liu, J.~Yan, Z.~Wang, and C.~Gong, ``An improved structured mesh generation method based on physics-informed neural networks,'' \emph{arXiv preprint arXiv:2210.09546}, 2022.

\bibitem{lu2021learning}
L.~Lu, P.~Jin, G.~Pang, Z.~Zhang, and G.~E. Karniadakis, ``Learning nonlinear operators via deeponet based on the universal approximation theorem of operators,'' \emph{Nature machine intelligence}, vol.~3, no.~3, pp. 218--229, 2021.

\bibitem{kovachki2023neural}
N.~Kovachki, Z.~Li, B.~Liu, K.~Azizzadenesheli, K.~Bhattacharya, A.~Stuart, and A.~Anandkumar, ``Neural operator: Learning maps between function spaces with applications to pdes,'' \emph{Journal of Machine Learning Research}, vol.~24, no.~89, pp. 1--97, 2023.

\bibitem{li2020fourier}
Z.~Li, N.~Kovachki, K.~Azizzadenesheli, B.~Liu, K.~Bhattacharya, A.~Stuart, and A.~Anandkumar, ``Fourier neural operator for parametric partial differential equations,'' \emph{arXiv preprint arXiv:2010.08895}, 2020.

\bibitem{tripura2023wavelet}
T.~Tripura and S.~Chakraborty, ``Wavelet neural operator for solving parametric partial differential equations in computational mechanics problems,'' \emph{Computer Methods in Applied Mechanics and Engineering}, vol. 404, p. 115783, 2023.

\bibitem{xiong2024koopman}
W.~Xiong, X.~Huang, Z.~Zhang, R.~Deng, P.~Sun, and Y.~Tian, ``Koopman neural operator as a mesh-free solver of non-linear partial differential equations,'' \emph{Journal of Computational Physics}, p. 113194, 2024.

\bibitem{cao2023lno}
Q.~Cao, S.~Goswami, and G.~E. Karniadakis, ``{LNO}: Laplace neural operator for solving differential equations,'' \emph{arXiv preprint arXiv:2303.10528}, 2023.

\bibitem{jin2022mionet}
P.~Jin, S.~Meng, and L.~Lu, ``{MIONet}: Learning multiple-input operators via tensor product,'' \emph{SIAM Journal on Scientific Computing}, vol.~44, no.~6, pp. A3490--A3514, 2022.

\bibitem{lu2022comprehensive}
L.~Lu, X.~Meng, S.~Cai, Z.~Mao, S.~Goswami, Z.~Zhang, and G.~E. Karniadakis, ``A comprehensive and fair comparison of two neural operators (with practical extensions) based on fair data,'' \emph{Computer Methods in Applied Mechanics and Engineering}, vol. 393, p. 114778, 2022.

\end{thebibliography}

\end{document}